\pdfoutput=1

\documentclass[11pt]{article}

\usepackage[final]{acl}

\usepackage{times}
\usepackage{latexsym}
\newcommand{\frameworkname}[1]{MOSAIC}

\definecolor{realcyan}{RGB}{0, 200, 200}

\usepackage[T1]{fontenc}

\usepackage[utf8]{inputenc}

\usepackage{microtype}

\usepackage{inconsolata}

\usepackage{graphicx}

\usepackage{amsmath}
\usepackage{booktabs} 
\usepackage{pdflscape}
\usepackage{longtable}
\usepackage{array}
\usepackage{multirow}
\usepackage{tabularx}
\usepackage{siunitx}
\usepackage{makecell}  
\usepackage[most]{tcolorbox}
\usepackage{listings}

\definecolor{mcolor}{RGB}{26, 54, 93}       
\definecolor{ocolor}{RGB}{0, 169, 166}      
\definecolor{scolor}{RGB}{207, 82, 48}      
\definecolor{acolor}{RGB}{228, 160, 37}     
\definecolor{icolor}{RGB}{0, 122, 83}       
\definecolor{ccolor}{RGB}{112, 47, 138}     

\definecolor{lightgray}{HTML}{E0E0E0}
\definecolor{lightblue}{HTML}{B3D9FF}
\definecolor{lightgreen}{HTML}{D2F8D2}
\definecolor{lightpurple}{HTML}{E6CCFF}

\newcommand{\frameworknamecolor}{%
\textbf{\textcolor{mcolor}{M}%
\textcolor{ocolor}{O}%
\textcolor{scolor}{S}%
\textcolor{acolor}{A}%
\textcolor{icolor}{I}%
\textcolor{ccolor}{C}}%
}

\title{\frameworknamecolor{}: Modeling Social AI for Content Dissemination and Regulation in Multi-Agent Simulations}

\author{
  Genglin Liu$^{1}$ \quad Vivian Le$^{1}$ \quad Salman Rahman$^{1}$ \quad \\ {\bf Elisa Kreiss$^{1}$ \quad Marzyeh Ghassemi$^{2}$ \quad Saadia Gabriel$^{1}$} \\
  $^{1}$ University of California, Los Angeles \quad $^{2}$ MIT CSAIL \\
  \texttt{{genglinliu}@cs.ucla.edu}
}

\begin{document}

\maketitle

\begin{abstract}
We present a novel, open-source social network simulation framework \frameworknamecolor~where generative language agents predict user behaviors such as liking, sharing, and flagging content. This simulation combines LLM agents with a directed social graph to analyze emergent deception behaviors and gain a better understanding of how users determine the veracity of online social content. By constructing user representations from diverse fine-grained actual user personas, our system enables multi-agent simulations that model content dissemination and engagement dynamics at scale. Within this framework, we evaluate three different content moderation strategies with simulated misinformation dissemination, and we find that they not only mitigate the spread of non-factual content but also increase user engagement. In addition, we analyze the trajectories of popular content in our simulations, and explore whether simulation agents' articulated reasoning for their social interactions truly aligns with their collective engagement patterns. We open-source our simulation software to encourage further research within AI and social sciences: \url{https://github.com/genglinliu/MOSAIC} 

\end{abstract}

\section{Introduction}

\begin{figure*}[htp]
    \centering
    \includegraphics[width=\textwidth]{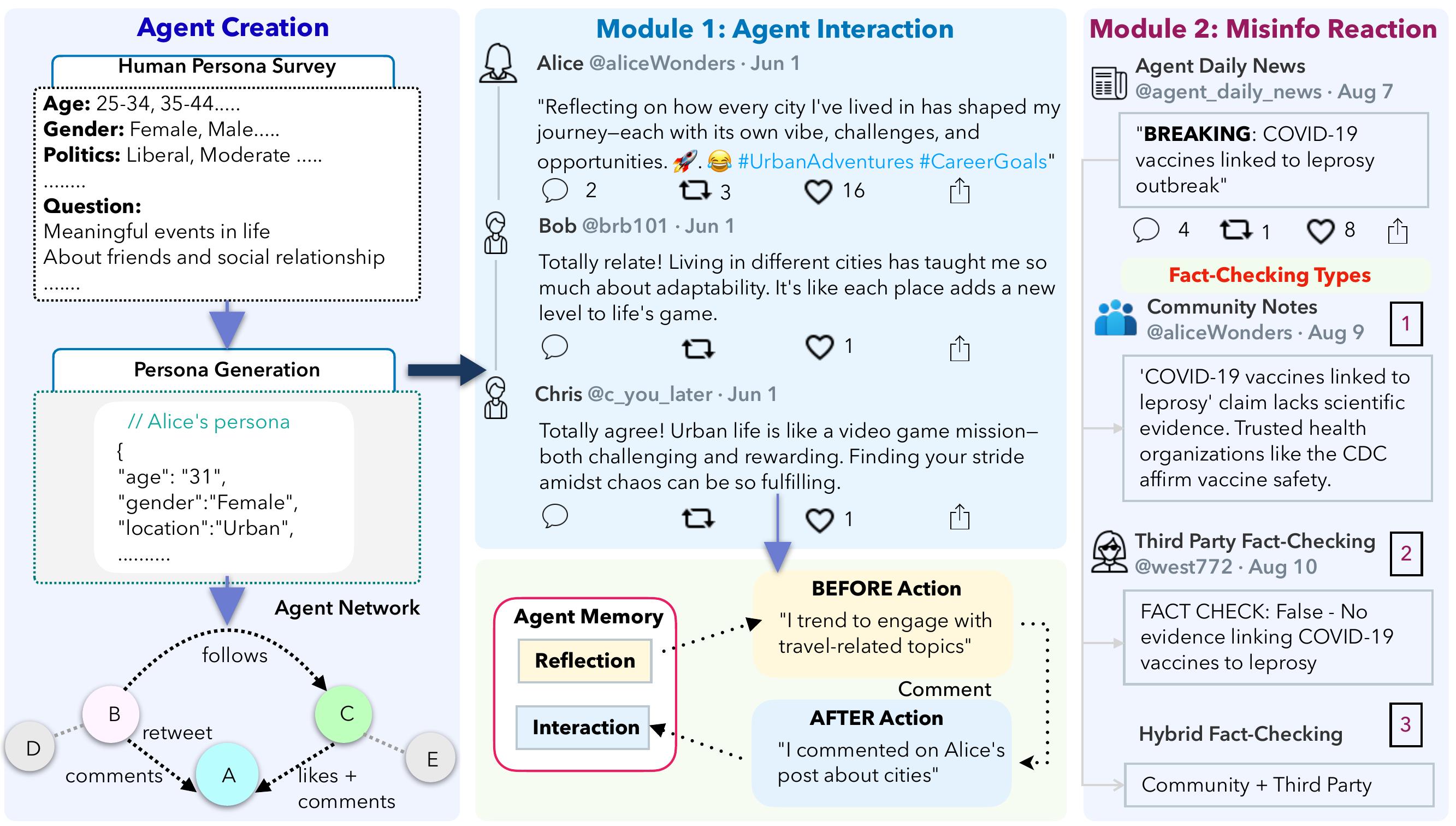}
    \caption{Overview of the \frameworknamecolor{}, a multi-agent social simulation framework where agents interact in an environment mimicking a social network, form dynamic memory-based behaviors, and respond to misinformation using community-based, third-party, or hybrid fact-checking mechanisms. Personas are replicated from human surveys or generated using synthetic distributions. Memories are retrieved before an agent takes certain actions, and are updated after certain events.}
    \label{fig:fig1}
\end{figure*}

In 2024, OpenAI reported that its platform was already being misused by covert influence operations to generate synthetic content diffused over social media \cite{disrupt}. These internet manipulators exploit the fact that social networks have become a fundamental part of modern life, shaping public discourse, influencing political opinions, and facilitating the rapid spread of unverified human- and AI-generated content \cite{aichner2021twenty, orben2022windows, cinelli2021echo}. While traditional social science methods such as surveys and observational studies have provided insights into human behavior, they often struggle to capture large-scale, emergent online interactions \cite{yu2021modeling, lorig2021agent}. Agent-based modeling (ABM) provides distinct advantages over survey methods in social science research since it can simulate dynamic interactions over time, and support examination of hypothetical or counterfactual scenarios with repeatable and controllable conditions \cite{bonabeau2002agent, epstein1999agent}.

Recent advances in foundation models have led to the emergence of social simulations with generative agents, where AI-powered users dynamically engage in social behaviors \cite{yang2024oasis, Gao2023S3SS, chen2024agentverse}. Unlike traditional survey methods or classical agent-based modeling, simulations driven by LLMs enable agents to interact with the environment and each other naturally through rich, human-like dialogue, closely mirroring authentic social behavior \cite{wang2024sotopia, zhou2023sotopia, park2022social}. In this work, we introduce \textbf{\frameworknamecolor, a novel multi-agent AI social network simulation that models content diffusion, user engagement patterns, and misinformation propagation.} 

Among different applications of social simulations, content moderation stands out as a pressing challenge due to the real-world harm caused by mis- or disinformation and online influence operations \citep{jhaver2023personalizing, young2022much}. Previous research has shown that false information not only spreads more rapidly and deeply than truthful content  \cite{vosoughi2018spread} but also alters public perception in ways that are difficult to reverse \cite{lewandowsky2012misinformation}. Addressing this issue requires effective content moderation strategies that can mitigate harm while preserving user engagement and freedom of expression. We embed three moderation strategies into our simulation environment:
\textbf{(1) community-based fact-checking} mimicking X and Meta's Community Notes, \textbf{(2) independent fact-checking}, and \textbf{(3)} a hybrid approach mixing \textbf{(1)} and \textbf{(2)}. We systematically evaluate the impact of these 3 content moderation strategies (along with a baseline of no fact-checking) on misinformation spread, moderation precision/recall, and user engagement dynamics.

Beyond moderation, understanding how certain content gains traction remains an open challenge. Online discourse is shaped by the dynamics of content diffusion, where some posts attract widespread engagement while others remain largely unseen. In our simulation, LLM-powered agents are equipped with memory, self-reflection, and explicit reasoning mechanisms, allowing them to explain their decisions and adapt their behavior over time. While our primary focus is on moderation, this extended perspective helps contextualize how misinformation and other content propagate in online interactions. To this end, our key contributions are:

\begin{itemize}
    \item We build a multi-agent social simulation that is shown to be high fidelity through validation against known social media behavioral phenomena and direct comparison with actual online users. Notably, we find that \textbf{generative agents are capable of accurately modeling individual engagement patterns, given fine-grained and realistic demographic portraits from user surveys} (\S\ref{sec:human-validation}).

    \item We conduct a comparative study of third-party, community-based, and hybrid fact-checking approaches, quantifying their effectiveness in mitigating misinformation while preserving engagement. We show that \textbf{misinformation doesn't spread as fast in an agent simulation as is commonly observed in human social media, and content moderation strategies can improve not only fact-checking but also engagement} (\S\ref{sec3.2}, \ref{sec3.3}).

    \item We explore how different content and network properties influence diffusion dynamics, offering insights into engagement patterns and how some content/users end up attracting more attention than others. We probe whether the agents' personas and content topics correlate with engagement. Surprisingly, we find that \textbf{agents’ individual verbose reasoning may not faithfully reflect their collective action patterns on a group level} (\S\ref{sec:content-virality}).
\end{itemize}

By bridging social science observations, game-theoretic modeling \cite{acemoglu2023model}, and LLM-driven modeling, our work demonstrates the potential of generative agent simulations as a tool for studying large-scale online behaviors.

\section{\frameworknamecolor{} Social Network Simulation}
\label{simulation}

\begin{figure}[ht]
    \centering
    \includegraphics[width=0.45\textwidth]{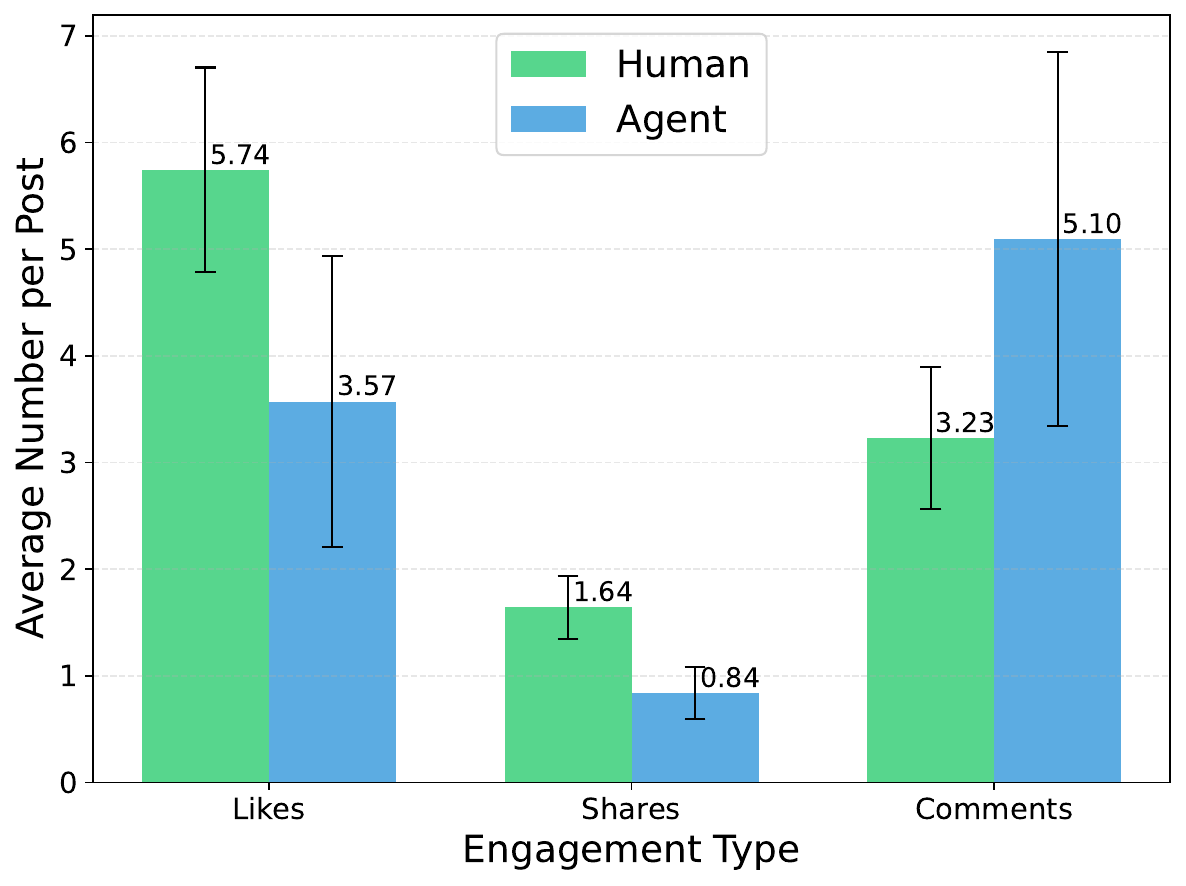}
    \caption{Average engagement received per post: Human vs. Agents. Our t-test validates that the difference in reaction patterns across the three engagement types are not statistically significant, suggesting that agents can simulate individual human reactions to social media feed realistically.}
    \label{fig:engagement_per_post}
\end{figure}


Our AI-driven social network simulates how content spreads, how users engage, and how misinformation propagates within a directed social graph. As illustrated in Fig.~\ref{fig:fig1}, at its core, the system simulates a dynamic environment where AI agents interact by following others, posting content, reacting (e.g., liking, sharing, commenting), and reporting misinformation. Each agent operates with a persona generated using a question set inspired by AgentBank \cite{park2024generative}. The main simulation system tracks the progression of time and the evolving state of the network. It is supported by several key components: a relational database that records all user interactions; a content manager that injects new posts into the network; an analytics module that monitors diffusion patterns and user behavior; and a fact-checking system that evaluates the performance of various content moderation strategies.

\paragraph{Simulated Network} We build a simulated social network environment inspired by platforms like X,\footnote{\url{https://x.com/}}, allowing AI-driven users to interact, post, and share content. The simulation includes a basic user class with attributes such as username, posts, followers, following, and reposts, mimicking the structure of real-world social media platforms. The network itself is defined by the follower-following relationships, creating a web of user interactions, represented by a directed graph $G = (N, E) $ where $N$ represents the set of user nodes, i.e., $N = \{n_1, n_2, \dots, n_k\}$, where $n_i$ is a user in the network. $E \subseteq N \times N$ represents the set of directed edges, i.e., $E = \{(n_i, n_j) \mid n_i \text{ follows } n_j \}$. Each edge $(n_i, n_j)$ signifies that user $n_i$ follows user $n_j$. 

\subsection{Simulation Flow}

The simulation begins with an initialization phase where the system loads experimental configurations, sets up the database (details in Appendix~\ref{appendix_database}), generates an initial user population (more details in Appendix~\ref{appendix_persona}), and establishes follow relationships. Agents are configured to operate under diverse behavioral traits, reflecting real-world variations in social media engagement. In all of our experiments, agents are driven by \texttt{gpt-4o} \cite{hurst2024gpt} as the foundation model backbone, unless otherwise specified. We do also implement an option to connect agents with open-weight models through SGLang \cite{zheng2024sglang} or vLLM's \cite{kwon2023efficient} inference engines.

At each time step, news content is introduced based on predefined parameters, with agents dynamically responding to their feeds. Agents can optionally generate posts according to their own interests. However, during certain controlled experiments, we configure them to only engage through reactions such as liking, sharing, commenting, or reporting misinformation. We describe a more general action space and more details of their decision-making process in Appendix~\ref{appendix_action_space}. The visibility of posts evolves based on engagement metrics, simulating algorithmic amplification effects. If fact-checking is enabled, agents incorporate moderation signals, e.g., they are prompted to pay more attention to potentially falsified content or misinformation. 
We discuss the content moderation simulation in more depth in Section \ref{sec3:content-moderation}. Throughout a simulation run, the system tracks key statistics, including content reach, user influence, and misinformation spread.
At the end of each simulation run, a post-hoc analysis is conducted to assess content diffusion dynamics, user engagement metrics, influence distribution, and the impact of fact-checking interventions. We also keep track of various network properties such as centrality\footnote{The degree to which an user is central to the network, having outsized or undersized influence.} and triadic closures,\footnote{A common social phenomenon where users with a mutual connection are more likely to connect to each other.} and perform homophily\footnote{Another common social phenomenon where similar users are more likely to connect to each other over dissimilar ones.} analysis to examine clustering patterns in user engagement (see Appendix~\ref{app:content_popularity_extended} for an extended technical description of how we define and compute these network metrics). Our focus on these three properties is motivated by a large body of prior work in network theory and the social sciences \cite{Rapoport1953SpreadOI, Granovetter1973TheSO, McPherson2001BirdsOA, Abebe2022OnTE,Chang2024LLMsGS}.

\subsection{Human Validation}
\label{sec:human-validation}

To validate the veracity of our simulation, we conducted a human study to compare the sharing patterns between humans and LLM agents. We recruited 204 participants via Prolific.\footnote{\url{https://www.prolific.com/}} More details of our human survey are provided in Appendix~\ref{app:human_study_details}.

\paragraph{Setup}
In the first phase of this replication study, we conducted a survey to collect demographic data (e.g., age, gender, religion, ethnicity, education level, language, residence, income, political stance) and personal values and behaviors (e.g., hobbies, residential history, social goals, meaningful life events, valued friendship traits, financial habits). Inspired by \citet{park2024generative}, we used this anonymized data to create individualized personas for 204 LLM-driven agents, each corresponding to a human participant.

In the second phase, both participants and their corresponding LLM agents were shown two curated social media snapshots containing 30 posts. 10\% of the articles are false news articles verified by an independent and non-partisan team of journalists from NewsGuard.\footnote{\url{https://www.newsguardtech.com/}} Study participants were instructed to respond to each post using a fixed set of actions (e.g., like, dislike, comment, share). The agents, guided solely by their assigned persona profiles, followed the same instructions. We then analyzed and compared engagement patterns between humans and agents, both overall and across demographic groups, to assess how well LLMs can emulate human social media behavior based on persona information alone.

\paragraph{Simulation/Human Reaction Alignment} 

We analyze the engagement behavior alignment between the human participants and the same number of persona-replicated AI agents, using independent two-sample t-tests for each engagement type. As illustrated in Fig.~\ref{fig:engagement_per_post}, no statistically significant differences emerged in likes (t = 1.33, p = 0.19) or comments (t = -1.05, p = 0.30), though humans gave slightly more likes (+2.17 per post), and agents posted more comments (+1.87 per post). A marginally significant difference was observed in shares (t = 2.11, p = 0.04), with humans sharing slightly more (+0.80 per post). These results indicate that persona-driven AI agents display engagement patterns that closely mirror those of humans, supporting the realism of our simulation. We also provide more details about the per-demographic engagement pattern alignment in Appendix~\ref{app:human_study_details}.

\begin{figure}[ht]
    \centering
    \includegraphics[width=0.45\textwidth]{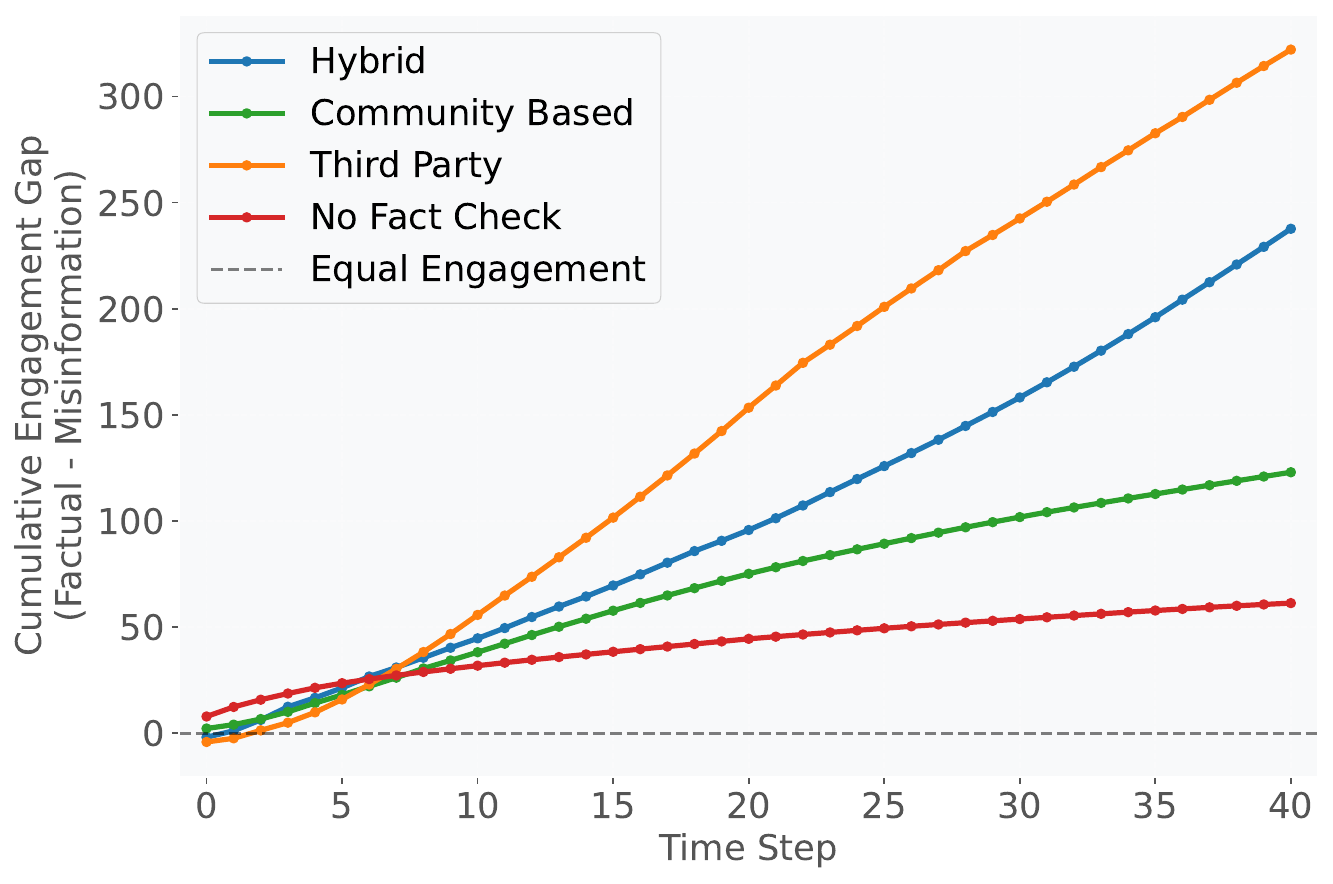}
    \caption{Effectiveness of content moderation approaches in promoting factual content, across models. Positive values: factual content receives more engagement. Negative values: misinformation receives more engagement.}
    \label{fig:cumulative_growth_gap}
\end{figure}

\section{Content Moderation in Simulated Social Environment}
\label{sec3:content-moderation}

\begin{table*}[ht]
    \centering
    \renewcommand{\arraystretch}{1.1}
    \resizebox{\textwidth}{!}{%
    \begin{tabular}{llccc|ccc|ccc|ccc}
        \toprule
        \bfseries Model & \bfseries Fact-Checking Method 
        & \multicolumn{3}{c}{\textbf{Post Statistics}} 
        & \multicolumn{3}{c}{\textbf{Factual Engagement (\# of)}} 
        & \multicolumn{3}{c}{\textbf{Misinfo Engagement (\# of)}} 
        & \multicolumn{3}{c}{\textbf{Fact-Checking Performance (\%)}} \\
        
        \cmidrule(lr){3-5}
        \cmidrule(lr){6-8}
        \cmidrule(lr){9-11}
        \cmidrule(lr){12-14}
        
        & 
        & \bfseries \# of Posts & \bfseries Factual & \bfseries Misinfo 
        & \bfseries Shares & \bfseries Likes & \bfseries Comments 
        & \bfseries Shares & \bfseries Likes & \bfseries Comments 
        & \bfseries Precision & \bfseries Recall & \bfseries F1 Score \\
        
        \midrule
        \texttt{deepseek-v3} & Unmoderated                  & 420 & 378 & 42 & 14  & 2478 & 3968 & 0  & 32  & 117 & 0.0   & 0.0   & 0.0   \\
        \texttt{deepseek-v3} & Third Party FC         & 420 & 378 & 42 & 4   & 5676 & 8809 & 0  & 114 & 622 & 50.0  & 50.0  & 50.0  \\
        \texttt{deepseek-v3} & Community FC           & 420 & 378 & 42 & 11  & 1987 & 7324 & 0  & 15  & 110 & 63.3  & 45.2  & 52.8  \\
        \rowcolor[HTML]{DFF0D8}
        \texttt{deepseek-v3} & Hybrid FC              & 420 & 378 & 42 & 3   & 3564 & 9822 & 0  & 42  & 353 & 100.0 & 45.2  & 62.3  \\

        \midrule
        \texttt{gpt-4o-2024-08-06} & Unmoderated                    & 500 & 450 & 50 & 84  & 69   & 193  & 2  & 1   & 9   & 0.0   & 0.0   & 0.0   \\
        \texttt{gpt-4o-2024-08-06} & Third Party (5p, offline)           & 450 & 405 & 45 & 752 & 2387 & 3800 & 31 & 88  & 255 & 21.9  & 63.6  & 32.6  \\
        \texttt{gpt-4o-2024-08-06} & Third Party (5p, online)            & 420 & 378 & 42 & 430 & 1883 & 3020 & 20 & 73  & 335 & 22.2  & 100.0 & 36.4  \\        
        \texttt{gpt-4o-2024-08-06} & Third Party (10p, online)           & 420 & 378 & 42 & 616 & 1917 & 3066 & 27 & 50  & 131 & 51.4  & 90.0  & 65.5  \\
        \texttt{gpt-4o-2024-08-06} & Community FC             & 490 & 441 & 49 & 80  & 334  & 731  & 3  & 7   & 25  & 45.8  & 44.9  & 45.4  \\
        \rowcolor[HTML]{DFF0D8}
        \texttt{gpt-4o-2024-08-06} & Hybrid FC                & 500 & 450 & 50 & 302 & 2115 & 4706 & 7  & 48  & 194 & 73.7  & 56.0  & 63.6  \\

        \midrule
        \texttt{claude-3.7-sonnet} & Unmoderated                    & 420 & 378 & 42 & 55  & 583  & 482  & 2  & 4   & 32  & 0.0   & 0.0   & 0.0   \\
        \texttt{claude-3.7-sonnet} & Third Party FC           & 368 & 332 & 36 & 571 & 7411 & 5526 & 8  & 61  & 280 & 0.0   & 0.0   & 0.0   \\
        \texttt{claude-3.7-sonnet} & Community FC             & 420 & 378 & 42 & 15  & 821  & 659  & 0  & 0   & 0   & 100.0 & 50.0  & 66.7  \\
        \rowcolor[HTML]{DFF0D8}
        \texttt{claude-3.7-sonnet} & Hybrid FC                & 420 & 378 & 42 & 284 & 8320 & 7684 & 0  & 2   & 10  & 100.0 & 81.0  & 89.5  \\

        \bottomrule
    \end{tabular}%
    }
   \caption{
Comparison of fact-checking (FC) strategies across \texttt{deepseek-v3}, \texttt{gpt-4o-2024-08-06}, and \texttt{claude-3.7-sonnet} under four settings: 
\textit{Unmoderated}, \textit{Community FC}, \textit{Third Party FC}, and \textit{Hybrid FC}. 
For \texttt{gpt-4o-2024-08-06}, third-party setups vary by number of posts reviewed per step (5 or 10) and whether web search is used (\textit{online} vs.\ \textit{offline}). In all other settings, third party is defaulted to checking 5 posts per time step, without web search.
Columns are grouped into (1) \textbf{Post Statistics}, (2) \textbf{Engagement Metrics} (shares, likes, comments for factual and misinformation), 
and (3) \textbf{Fact-Checking Performance} (precision, recall, F1 score on misinformation).
}
    \label{tab:fact_checking_engagement_aligned}
\end{table*}

\begin{figure*}[ht]
    \centering
    \includegraphics[width=\textwidth]{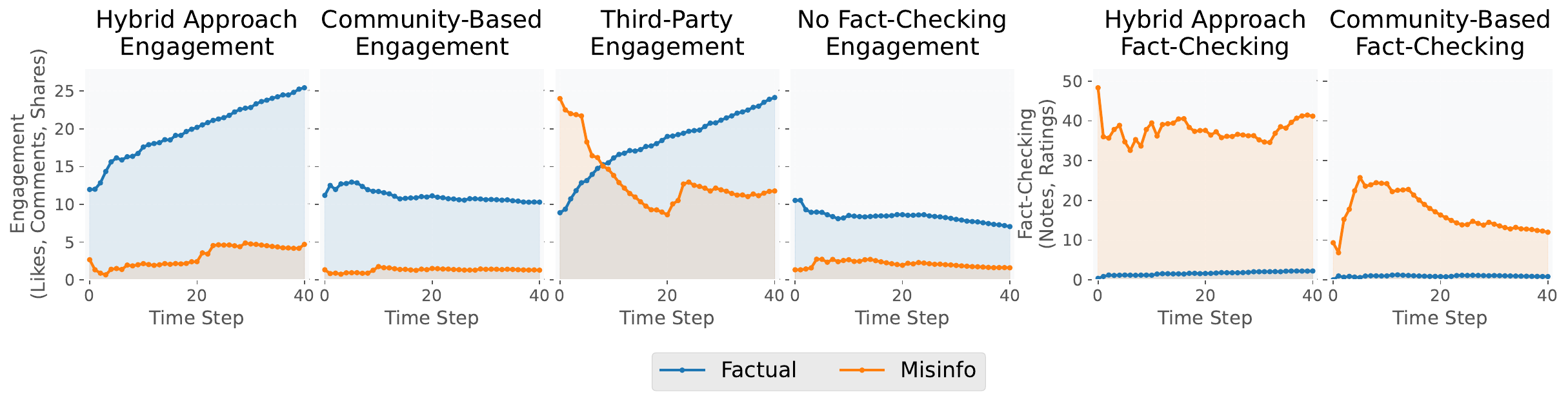}
    \caption{A consolidated view of content engagement and fact-checking metrics, averaged across all models used in our experiments. The first four panels display engagement metrics—specifically, the sum of likes, comments, and shares—under each fact-checking condition. The last two panels show fact-checking metrics, which combine both the number of community notes and note ratings, for the two methods where these are applicable (Hybrid and Community-Based). All values represent the average behavior across models, providing a holistic summary of the system’s dynamics under each experimental setting.}
    \label{fig:total_interactions}
\end{figure*}

We conduct a series of experiments using our multi-agent simulation framework to investigate the effects of different fact-checking strategies on the spread of both factual and misinformation content. Our findings reveal key differences between LLM-driven social simulations and human social networks in how misinformation propagates.

\subsection{Setup}

\paragraph{Data Sources}

We obtained a data license from NewsGuard to access proprietary information on widespread misinformation narratives tracked by their independent team of journalists. We collected 1,353 examples of false news from their database with release dates up to December 19th, 2024.\footnote{We removed non-English articles and de-deduplicated.}

To collect factual news articles, we utilize a news aggregation API\footnote{\url{https://newsapi.org/}} to retrieve articles published daily in legitimate sources from January 31 to February 28, 2025. The system queries the API for all available topics, prioritizing non-political popular articles in English. For each date in the range, we extract key information from the retrieved articles, including their title, description, and main content. Using the NewsAPI, we scraped a total of 2470 news articles from major media outlets. 


\paragraph{Environment Initialization} Our simulations involve agentic users interacting with news posts under four different fact-checking conditions: (1) No Fact-Checking, (2) Community-Based Fact-Checking, (3) Third-Party Fact-Checking with an independent LLM that uses its own parametric knowledge, and (4) Hybrid Fact-Checking, which integrates both community-based and third-party verification mechanisms. The simulations start with 50 agents and span over 40 time steps, with agents making interaction decisions based on the perceived veracity of posts and the presence of content moderation. At each new step, we randomly introduce up to 2 more agents into the environment to simulate the regular user growth of the social media platform. We analyze both the overall engagement with posts and the effectiveness of fact-checking strategies in suppressing misinformation.

\paragraph{Fact-Checking Settings} The action space of agents varies across different fact-checking conditions, reflecting different levels of scrutiny and intervention in their social media interactions. In the \textbf{\textit{no fact-checking}} setting, agents interact freely with the feed, engaging with posts based solely on their interests and beliefs. They can like, share, comment, or ignore posts without any explicit instructions to assess the accuracy of the content. Under this setting, a post is considered ``fact-checked" when it has at least one community note with a helpful rating. In the \textbf{\textit{third-party fact-checking}} condition, the action space remains the same, but the environment implicitly assumes the presence of external fact-checkers who may influence the visibility or credibility of posts. However, the agents themselves do not perform any direct verification. In contrast, the \textbf{\textit{community fact-checking}} setting expands the action space by allowing agents to add community notes to posts they deem misleading or in need of additional context, as well as rate existing community notes as either helpful or unhelpful. This introduces a participatory element, encouraging agents to contribute to a crowdsourced verification system. Finally, the \textbf{\textit{hybrid fact-checking}} condition combines elements from both third-party and community-driven verification. Agents can engage with posts as in previous settings while also considering official fact-checks alongside community notes, contributing their own notes and rating those written by others. Across all conditions, agents must select from predefined valid actions, ensuring consistency in response formats. Additionally, when reasoning is enabled, agents are required to justify their interactions by providing a brief explanation for each chosen action, further enhancing the interpretability of their behavior.

\paragraph{Fact Checker LLM} The third-party fact checker is represented by an automated content verification system designed to identify and address misinformation on a social platform. It works by prioritizing posts for review based on engagement metrics (likes, shares, comments), news classification, and user flags, with special priority given to content that has received community notes in hybrid fact-checking scenarios. The system leverages an independent LLM instance: \texttt{gpt-4o} (without web browsing capability) and \texttt{perplexity-sonar-pro}\footnote{Perplexity AI: \url{https://sonar.perplexity.ai/}} (with web browsing) in our experiments. The system categorizes posts as "true," "false," or "unverified," each accompanied by a verdict explanation, confidence score, and supporting evidence sources. When posts are deemed false with high confidence ($\ge 0.9$ in third-party fact-checking mode, or $\ge 0.7$ in hybrid mode with community notes), the system automatically takes them down and records the justification. All fact-check results are stored in the database that maintains an audit trail of verdicts alongside ground truth data when available.

\paragraph{Network Initialization} We initialize a scale-free network of LLM-powered agents interacting within a directed social graph, using a Barabási–Albert model \cite{barabasi1999emergence}. Misinformation and factual content are injected into the system at controlled rates. Each moderation strategy is implemented in a separate experiment, allowing for comparative analysis. We share more details about the experiment configurations in Appendix~\ref{appendix_exp_config}.

\subsection{False News Does Not Spread Faster than Real News with Simulation Agents}
\label{sec3.2}

A key insight from our simulation contradicts established results from human social networks: false news does not spread faster than real news. Prior studies on human social behavior have consistently demonstrated that misinformation propagates more rapidly and deeply than factual content \citep{vosoughi2018spread, zhao2020fake}. However, in our agent-based simulation, engagement (particularly with sharing) with misinformation does not surpass that of factual news. 

As shown in Fig.~\ref{fig:total_interactions}, factual news maintains consistently higher engagement levels than misinformation across all four settings. The gap is most pronounced under the Third-Party and Hybrid fact-checking conditions, where factual interactions climb steadily while misinfo remains low. Notably, in the No Fact-Checking scenario, false news still fails to gain a foothold, suggesting that these agents are inherently less likely to propagate unverified or misleading posts. We believe this behavior arises from two factors. First, the agents may rely on their internal confidence to guide sharing decisions, so low‐confidence (often false) content is passed over more often. And second, commercial LLMs have been post-trained with safety and helpfulness objectives. This training bias would make our agents less inclined to produce or circulate uncertain or harmful content.

This result highlights a key difference between LLM agents and human users. Human networks amplify sensational or controversial content, but our agents default to a more conservative sharing policy. Future work can explore more sophisticated modeling of cognitive biases or varied trust profiles to better mimic human spread patterns.




\subsection{Content Moderation Improves Both Fact-Checking and Engagement}
\label{sec3.3}

While political misinformation does not spread faster than factual news in our simulations, overall engagement remains low without any fact-checking. Fig.~\ref{fig:total_interactions} shows that, in the No Fact-Checking baseline condition, agents interact very little with both true and false posts. We hypothesize that LLM-driven agents, when uncertain about a post’s veracity, choose to disengage rather than risk amplifying misleading content.

Surprisingly, with the community-based, third-party, and hybrid fact-checking, we find that each not only suppresses misinformation (Tab.~\ref{tab:fact_checking_engagement_aligned}) but also increases engagement with factual content. In particular, Third-Party verification produces the largest cumulative advantage for real news, achieving $\sim$325 interactions by the final time step (Fig.~\ref{fig:cumulative_growth_gap}). Even the unmoderated baseline shows a modest upward trend for factual posts, hinting at an inherent bias toward accurate information in our network.

The steadily widening gaps between the third-party/hybrid approaches indicate that certain content moderation generates compounding benefits for factual engagement over time. Among the simulated strategies, fact-checking by third parties proves most effective at building a healthy information ecosystem—an effect that aligns with human studies showing increased trust in unflagged content \cite{pennycook2020implied}. All curves in Fig.~\ref{fig:total_interactions} represent metrics averaged across our three LLM agents (DeepSeek-V3, GPT-4o, and Claude-3.7-Sonnet). Interaction counts, including the number of shares, likes, comments, and notes, are aggregated at each time step to highlight overall platform dynamics under each moderation strategy. This high-level view illustrates consistent patterns in how content moderation shapes engagement. For a detailed, model-by-model breakdown and finer-grained metric analysis, please see Appendix~\ref{app:engagement_analysis_detailed}.

\subsection{Fact-Checking Performance: Hybrid Shows Superior Balance Across Models}

We evaluate fact-checking effectiveness using precision, recall, and F1 score (See Tab.~\ref{tab:fact_checking_engagement_aligned}) across all three models—DeepSeek-V3 \cite{liu2024deepseek}, GPT-4o \cite{hurst2024gpt}, and Claude-3.7-Sonnet \cite{claude37}. We find that LLM-based fact-checkers consistently perform best in the \textit{Hybrid Fact-Checking} setting. Claude outperforms GPT-4o and DeepSeek, achieving a F1 score of 0.895, with perfect precision (1.0) and high recall (0.810).  

\paragraph{Third Party Checking Ablation} For GPT-4o, we conduct an ablation over different third-party fact-checking configurations. Increasing the number of posts reviewed per step and enabling web search both lead to notable improvements in recall and overall F1 score. The 5-post offline setting yields a modest F1 score of 32.6\%, with low precision (21.9\%) and moderate recall (63.6\%). Switching to the 5-post online setup boosts recall to 100\%, indicating that the system catches all misinformation posts in this condition, though precision remains low (22.2\%), resulting in a slightly improved F1 score of 36.4\%. The best third-party performance is observed in the 10-post online setup, which balances both precision (51.4\%) and recall (90.0\%), achieving an F1 score of 65.5\%. These results suggest that increasing review coverage and leveraging web search significantly enhance the effectiveness of third-party fact-checking.

\subsection{What Drives Users/Content Popularity?}
\label{sec:content-virality}


\begin{figure}[ht]
    \centering
    \includegraphics[width=0.5\textwidth]{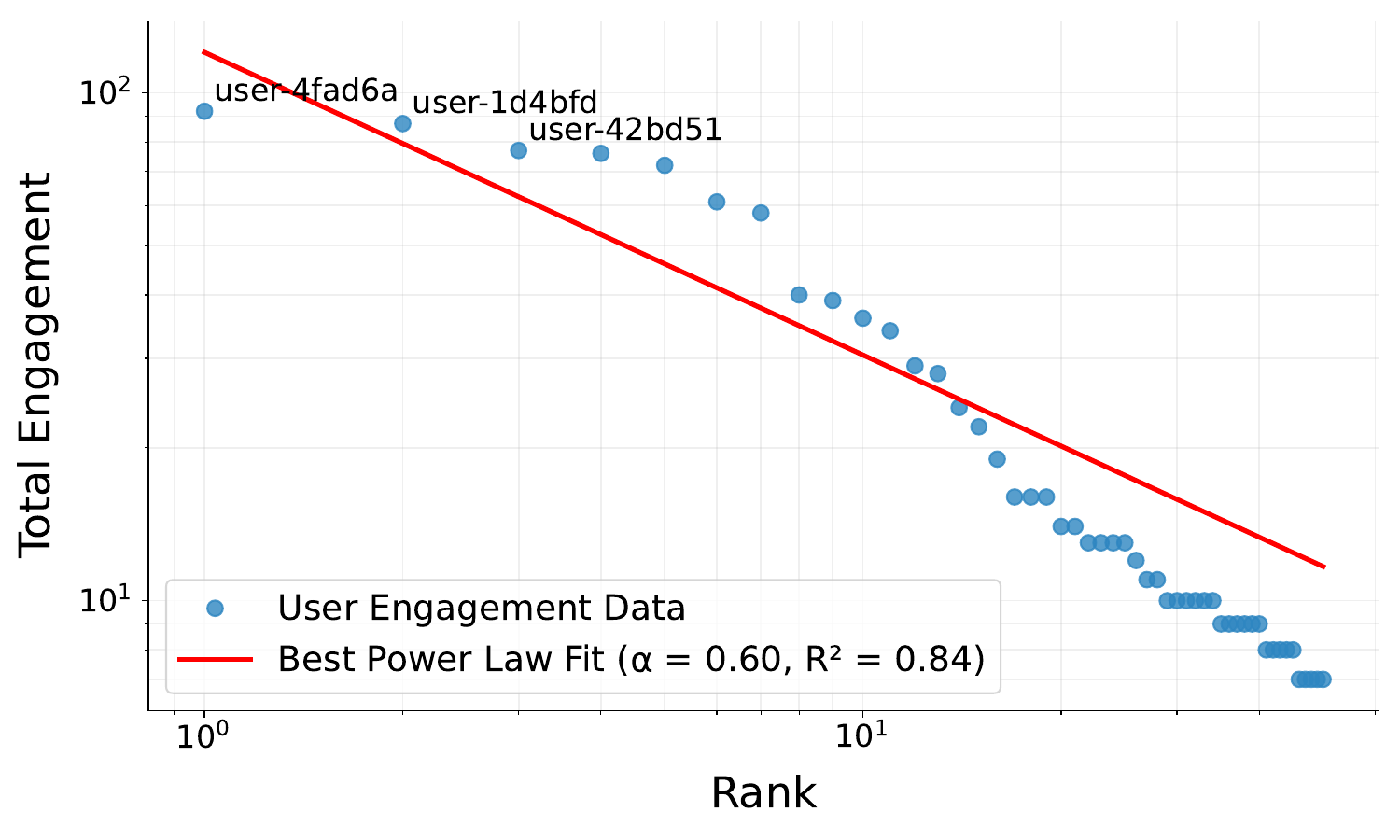}
    \caption{User engagement's best power-law fit. Engagement is defined as the sum of reposts, likes, and comments received by the user.}
    \label{fig:user_engagement_best_powerlaw_fit}
\end{figure}

Understanding what makes some content more popular is key to modeling engagement and intervention. In this standalone study, we simulate 161 agents over 4,249 posts of news and user-generated content, and let agents both react to and generate content based on personas, engagement cues, and memory-based decisions.

\begin{table}[htbp]
    \centering
    \caption{Chi-square Test Results for Differences in Engagement Based on Demographic Attributes}
    \label{tab:chi_square_results}
    \resizebox{\columnwidth}{!}{%
    \begin{tabular}{lcccc}
        \toprule
        \textbf{Attribute} & \textbf{Chi-square} & \textbf{p-value} & \textbf{Cramer's V} & \textbf{Effect Size} \\
        \midrule
        Age Group & 1.632 & 0.652 & 0.128 & Small \\
        Gender & 0.653 & 0.721 & 0.081 & Negligible \\
        Activity & 5.030 & 0.412 & 0.224 & Small \\
        Hobby & 9.101 & 0.246 & 0.302 & Medium \\
        Ethnicity & 10.187 & 0.070 & 0.319 & Medium \\
        Income Level & 4.373 & 0.358 & 0.209 & Small \\
        Political Affiliation & 2.515 & 0.642 & 0.159 & Small \\
        Primary Goal & 8.064 & 0.089 & 0.284 & Small \\
        \bottomrule
    \end{tabular}%
    }
\end{table}

\paragraph{Findings}
Engagement follows a heavy-tailed power law ($R^2=0.84$, $\alpha=0.60$; Figure~\ref{fig:user_engagement_best_powerlaw_fit}), even when user persona and the network graph are initialized randomly. Few users attract most interactions despite a lower exponent than real networks \citep{muchnik2013origins, bild2015aggregate}. Most persona demographic attributes show no significant effect on popularity (Tab.~\ref{tab:chi_square_results}), even though out of 8 attributes, ethnicity and hobby have medium effect sizes (Cramer's $V=0.319,0.302$) \cite{Cramr1946MathematicalMO}. Topic clusters via BERTopic on \texttt{all-MiniLM-L6-v2} \cite{grootendorst2022bertopic,reimers2019sentence} embeddings reveal no engagement differences (ANOVA $F=0.614$, $p=0.84$).

Finally, we examine agents' own reasoning traces to better understand engagement behaviors. Interestingly, agent reasoning traces show that sentiment and motivations do not always align with action types (e.g., positive sentiment in follows, likes, shares; quality or misinformation concerns in flags). As shown in Tab.~\ref{tab:agent_reasoning_summary}, 13.5\% of “flags” are coded as positive even though flag reasons are exclusively negative (information value, misinformation). The verbal reasoning for the “unfollow” actions is 20\% positive, though it is intuitively a negative reaction towards other users. “ignore” omits any neutral reasoning and has a 77.8\% positive rate in reasoning, which also makes their verbal explanations inconsistent with the actual action. Following these inconclusive analyses, we speculate that our recency-and-follow-based feed ranking creates preferential attachment: once followed, a user’s posts gain visibility, reinforcing their popularity. We provide further analyses in Appendix~\ref{app:content_popularity_extended}.

\section{Background}



\paragraph{LLM-Driven Social Simulations.}

LLMs have transformed agent-based modeling by enabling context-aware, generative behaviors. Early simulations such as Schelling’s segregation model \cite{schelling1971dynamic}, Sugarscape \cite{epstein1996growing}, and NetLogo-based environments \cite{Wilensky1999NetLogo1} rely on static heuristics; recent systems \cite{park2023generative, chen2024agentverse, piao2025agentsociety, chirper} showcase agents with lifelike interactions and social dynamics. However, LLM-driven agents still face challenges like inconsistency and limited long-term reasoning. Our work addresses this by incorporating structured constraints and iterative feedback to enhance reliability for social science research.

\paragraph{Misinformation and Fact-Checking.}

False information often spreads more rapidly than truth due to emotional appeal and engagement-driven algorithms \cite{vosoughi2018spread,Pennycook2021ThePO,solovev2022moral}. Existing responses—third-party fact-checking \cite{googleIndia,googleAus}, algorithmic detection, and crowdsourced moderation like Community Notes\footnote{\url{https://communitynotes.x.com/guide/en/}}—each face limitations in scalability, accuracy, or bias \cite{zannettou2019web,panizza2023online}. We use simulations to evaluate these approaches in controlled settings, comparing their effectiveness and exploring hybrid strategies.

\paragraph{Simulations for Governance and Policy.}

Simulations have long supported decision-making in fields like epidemiology and public policy \cite{currie2020simulation,axtell2022agent,qu2024performance, cai2025simulation}. In the context of social media governance, LLM-driven simulations offer a novel testbed for assessing content moderation and algorithmic interventions before deployment \cite{charalabidis2011enhancing,landau2024challenging, qiu2025can, gu2025large}. Our framework enables scalable experimentation with regulatory strategies, contributing to ongoing efforts in algorithmic auditing and platform accountability.

\section{Conclusion}

Our study introduces a novel generative multi-agent simulation to model content diffusion, engagement, and misinformation dynamics in social networks. Our proposed moderation, combining community-based and independent fact-checking, balances misinformation reduction and user engagement. Notably, LLM agents tend to avoid unverified content, likely due to safety training, and misinformation did not spread faster than factual news, unlike in human studies. Engagement followed a power-law distribution, with few users driving most activity. However, user attributes and content topics were weak predictors, highlighting the complexity of online ecosystems. Agent reasoning showed a gap between stated motivations and actual behavior, suggesting future work on the faithfulness of agentic reasoning.

\section*{Limitations}

Our findings are subject to several limitations, particularly in the scale of our experiments. First, the limited number of human participants, especially from minority demographic groups, restricts the statistical power of our conclusions. Expanding participant diversity would enable a more robust analysis of how alignment between real and simulated social interaction patterns varies across demographics. 

Second, our content moderation experiments were conducted at a relatively small scale, which may have constrained the emergence of complex behaviors. Running these experiments at a larger scale could uncover additional dynamics not captured in the present study. Moreover, the simulation platform itself simplifies several aspects of real-world social media. For example, feed ranking is based primarily on recency and follow edges, rather than incorporating sophisticated recommender algorithms. This design choice was intentional to isolate the causal effects of content and moderation strategies, but it may limit the external validity of our findings. Future work could integrate recommender-driven dynamics while maintaining interpretability. 

Finally, we observe a gap between agents’ explicit explanations for their actions and the collective reaction patterns that emerge in the system. The root causes of this misalignment remain unclear and warrant further investigation, potentially involving a deeper analysis of agent modeling assumptions or social influence mechanisms.

\section*{Ethical Considerations}

This work raises several ethical considerations. First, our simulations incorporate misinformation narratives from the proprietary NewsGuard database. While these examples were necessary for experimental validity, they were used strictly for research purposes and are presented with appropriate warnings to prevent accidental amplification. We do not reproduce or promote harmful claims beyond what is needed for reproducibility. 

Second, our human study with 204 participants was conducted under an approved protocol with informed consent. Participants were compensated fairly through Prolific. Sensitive demographic attributes were anonymized and used only in aggregate to generate agent personas. No personally identifying information is reported.

Third, our findings highlight that current LLMs exhibit safety alignment that reduces engagement with misinformation. While this property is beneficial in our study, it may not reflect real human susceptibility, and thus should not be interpreted as predictive of real-world behavior. We caution against overgeneralizing from simulation to deployment.

Finally, by releasing MOSAIC as an open-source framework, we aim to support responsible research in AI safety and computational social science. We encourage careful use of this tool, especially in studies that model or intervene in sensitive online behaviors, and emphasize that simulations should complement rather than replace empirical studies with human subjects.

\section*{Acknowledgment}
\label{sec:ack}

This work was supported in part by research credits generously provided by OpenAI. In addition, the authors would like to thank Jacob Andreas for his valuable feedback and insightful suggestions.

\bibliography{custom} 

\appendix

\section{Extended Discussion on Content Popularity}
\label{app:content_popularity_extended}

\subsection{Network Properties: Centrality and Clustering Metrics}

We analyze the structural role of each agent in the directed follow‐network $G=(V,E)$ with $N=|V|$ nodes by computing degree, betweenness, closeness, and eigenvector centralities, as well as the local clustering coefficient and global transitivity.

Degree centrality quantifies how many direct connections a node has relative to the maximum possible. For node $v$ with total degree $\deg(v)=k_v$ (in‐ plus out‐degree), we define
\[
C_D(v)=\frac{k_v}{N-1},
\]
and compute the average degree centrality as
\[
\overline{C_D}=\frac{1}{N}\sum_{v\in V} C_D(v).
\]

Betweenness centrality measures the fraction of shortest directed paths between all ordered pairs $(s,t)$ that pass through $v$. Let $\sigma_{st}$ be the number of shortest paths from $s$ to $t$, and $\sigma_{st}(v)$ those that pass through $v$. Then
\[
C_B(v)=\sum_{\substack{s,t\in V\\s\neq t\neq v}}\frac{\sigma_{st}(v)}{\sigma_{st}},
\]
and its normalized average is
\[
\overline{C_B}
=\frac{1}{N\,(N-1)\,(N-2)}\sum_{v\in V}C_B(v).
\]

Closeness centrality reflects how near a node is to all others based on shortest‐path distances $d(v,u)$. We set
\[
C_C(v)=\frac{N-1}{\sum_{u\in V\setminus\{v\}}d(v,u)},
\]
and average as
\[
\overline{C_C}=\frac{1}{N}\sum_{v\in V}C_C(v).
\]

Eigenvector centrality assigns importance proportional to the centrality of a node’s neighbors. If $\mathbf{A}$ is the adjacency matrix of $G$, we solve
\[
\mathbf{A}\,\mathbf{x}=\lambda_{\max}\,\mathbf{x},
\]
and take $C_E(v)=x_v$, with average
\[
\overline{C_E}=\frac{1}{N}\sum_{v\in V}x_v.
\]

The local clustering coefficient of node $v$ measures the density of edges among its $k_v$ neighbors. If $e_v$ is the number of edges between neighbors, then
\[
C(v)=\frac{2\,e_v}{k_v(k_v-1)},
\]
and the network‐wide clustering is
\[
\overline{C}=\frac{1}{N}\sum_{v\in V}C(v).
\]

Finally, transitivity (global clustering) is the ratio of closed triplets (triangles) to all connected triplets:
\[
T=\frac{\sum_{v\in V}2\,e_v}{\sum_{v\in V}k_v(k_v-1)}.
\]

In these definitions, $d(v,u)$ denotes the shortest directed‐path length from $v$ to $u$, $\sigma_{st}$ the total number of shortest paths from $s$ to $t$, and $\sigma_{st}(v)$ those passing through $v$. All sums run over $V$ unless noted otherwise.

In our simulated community of 161 agents engaging with 4,249 pieces of content, the resulting follow‐network comprises 358 directed ties, indicating a modest level of connectivity driven by personas and memory‐based decisions. An average degree centrality of 0.1105 shows that each agent follows roughly 11\% of the population, while the low average betweenness centrality (0.0128) suggests that few agents act as indispensable bridges. With an average closeness centrality of 0.1367, most agents remain just a few steps apart, supporting rapid information flow, and the modest eigenvector centrality (0.0433) reveals that influence is fairly distributed rather than monopolized by a handful of hubs. Finally, a clustering coefficient of 0.205 and network transitivity of 0.189 point to moderate local grouping without excessive fragmentation—together painting a picture of a network that balances cohesion and reach in propagating popular content.

\subsection{Power-Law Distribution of User Popularity}

First, we define the popularity of users as a sum of the number of followers, number of likes, shares, and comments received. We collected the top 50 users and plotted their popularity (as measured by the sum of engagement received by them) from highest to lowest in Fig.~\ref{fig:top50_users}. We observe a power-law distribution of user influence. We have $ f(x) = 120x^{-0.6}$ as the best-fitting power-law approximation of our sampled data, as shown in Fig.~\ref{fig:user_engagement_best_powerlaw_fit}. With $\alpha = 0.60$, our regression line has an $R^2 = 0.84$, suggesting a strong fit to our user engagement data and that our social system follows a typical power law distribution where a few users generate most of the engagement. Existing analysis on real-world social networks suggests that this power-law exponent usually ranges from 1.5-2.5, depending on the specific context \citep{muchnik2013origins, bild2015aggregate}. Our best-fit exponent is lower than these reported numbers, but it still illustrates a clear trend that a minority of users/content collect most of the engagement, while the majority of them do not contribute nearly as much.

In the rest of this section, we explore potential reasons why this distribution emerges, and through a series of analyses leveraging our simulated environment, we reveal the unpredictability of influence or popularity in online social networks. More fundamentally, we argue that perhaps LLM-driven agents have a tendency to simply copy the decisions of agents who act before them. This results in the preferential attachment and, as a natural consequence, establishes the power-law distribution of engagement patterns. Such a pattern does not necessarily stem from anything else, such as the user's profile details or the content they post about. And even their own "inner reasoning" might not reveal their true decision-making, which invites further investigation into the authenticity of LLM agents' self-expressed reasoning traces. 

\subsection{Persona Attributes Don't Correlate with Engagement}

We analyzed user engagement by comparing the top 50 most engaged users (highest number of followers, likes, shares, comments, etc.) with the bottom 50 least engaged users across several attributes. The Chi-square test results summarized in Tab.~\ref{tab:chi_square_results} indicate that there are no statistically significant differences in the distributions of age group, gender, activity type, hobby, ethnicity, income level, political affiliation, or primary goal between the two groups. Although some attributes, such as ethnicity and hobby, exhibited medium effect sizes (Cramer's V \cite{Cramr1946MathematicalMO} of 0.319 and 0.302, respectively), their associated p-values did not reach conventional levels of statistical significance. This suggests that the attributes examined do not notably influence the level of user engagement.


Notably, we did not include personas resembling real-world public figures or celebrities, whose presence might have substantially influenced content popularity. Our findings thus suggest that, when personas are initialized randomly, some users naturally attract significantly more attention and engagement, independent of the specific attributes assigned during their initialization. This underscores the inherent variability and unpredictability of user engagement in social platforms.

\subsection{Do Content Topics Matter?}
Our analysis aimed to directly investigate the correlation between content topics and user engagement. To accomplish this, we first computed an engagement score for each post by summing its likes, shares, and comments. We then cleaned and preprocessed the textual content of the posts to ensure accurate topic modeling.

For topic extraction, we employed a unified topic model based on BERTopic \cite{grootendorst2022bertopic}, utilizing sentence embeddings from the SentenceTransformer model \texttt{all-MiniLM-L6-v2} \cite{reimers2019sentence}.BERTopic was chosen due to its effectiveness in capturing nuanced semantic relationships within short text content. By fitting a single topic model to all posts, we ensured consistency and comparability across the identified topics.

Following the topic assignment, we conducted a detailed statistical analysis. Engagement metrics—including mean, median, and standard deviation for likes, shares, comments, and overall engagement scores—were calculated for each topic.
To statistically assess whether variations in engagement across topics were significant, we performed an ANOVA (Analysis of Variance).

The key statistical finding from the analysis was an ANOVA result yielding an F-statistic of 0.614 and a p-value of 0.84. This indicates \textbf{no statistically significant relationship between the topics and overall engagement levels}. In other words, statistically, the topic of a post alone does not reliably predict its engagement level.

\subsection{Clues from Agents' Own Reasoning Traces and Recommender System}

The lack of clear correlation between user profiles, content topics, or temporal properties, and engagement patterns suggested that maybe the way we present the feed to the agents influences what content ends up being popular. Here, we discuss our feed prioritization algorithm. Our simulation does not employ a sophisticated recommender system. Our feed prioritization in the simulation relies primarily on recency and existing follow relationships, rather than explicit engagement metrics such as likes or shares. Posts, regardless of whether they're from followed or non-followed users, are generally ordered based on creation time, ensuring that newer content receives greater visibility. However, content from followed users gains additional prioritized exposure due to dedicated allocations in the feed. This structure might create a follower-based feedback loop: when User B follows User A, A’s posts consistently appear in B’s feed, enhancing A’s opportunities for engagement through likes, comments, and shares. Higher engagement subsequently boosts A’s visibility to other users who view these interactions, increasing the likelihood of additional follows and further amplifying this cycle.

\paragraph{Agent's Reasoning Pattern}

We extract and analyze agent reasoning across several dimensions, including sentiment, motivation, entity and concept extraction, and word-frequency analysis. The analysis specifically focused on identifying patterns related to different engagement actions (such as likes, comments, and shares), exploring how post content and user backgrounds influenced reasoning, and examining common linguistic trends. 

The analysis of agent reasoning reveals patterns in how agents engage with content and users on social media. As shown in Fig.~\ref{tab:agent_reasoning_summary}, agents demonstrate clear and distinct emotional sentiment patterns associated with different types of actions. Positive-dominant actions such as following users (99\% positive sentiment), commenting (97\%), liking posts (92\%), and sharing content (92\%) indicate that agents predominantly perceive their interactions as constructive contributions. Conversely, negative sentiment predominantly characterizes actions like flagging posts (71\% negative) and unfollowing users (40\% negative), reflecting agents' use of these interactions primarily for expressing disapproval or concern.

Further examining motivational reasoning, agents apply distinct frameworks depending on the nature of their engagement. Content evaluation actions, such as flagging posts, are predominantly motivated by information quality assessments (49\%) and concerns regarding misinformation (22\%). Sharing decisions primarily reflect agreement with content (46\%). In contrast, relationship-building actions show different motivations: liking is heavily driven by social connection potential (34\%), commenting balances agreement (29\%) and social connection (28\%), and following users reflects diverse personal interests (27\%).

Vocabulary analysis further emphasizes these distinctions, revealing specialized linguistic patterns for each type of engagement. Flagging content uses specific moderation-related language such as "misinformation," "harmful," and "credible," whereas community-oriented engagements like sharing, liking, and commenting frequently reference concepts like "community," "support," and "alignment." The "follow-user" action highlights terms related to content curation and long-term value, including "consistently," "valuable," and "insights."

Interestingly, despite these detailed reasoning frameworks, \textbf{a low alignment (21.4\%) between post sentiment and agent reasoning indicates that agents' explicit justifications may not fully reflect the underlying factors driving engagement. Instead, engagement decisions appear largely guided by personal values alignment}, information quality assessments, community-building potential, and personal relevance rather than simple emotional resonance with content.

These insights also highlight a notable contradiction with prior analyses, which showed no significant correlation between user demographics or content topics and overall engagement popularity. \textbf{While agents clearly articulate their engagement motivations in terms of specific frameworks (values alignment, informational quality, social connection), these explanations alone do not robustly predict broad engagement patterns.} This paradox suggests that engagement is heavily individualized, contextual, and possibly influenced by network effects—such as who posts content, existing social validation, or content placement within social feeds—factors not fully captured by demographic or topical categorizations alone.

In essence, the analysis confirms that agents employ reasoning structures tailored to the type of engagement but reveals that actual engagement outcomes are influenced by nuanced individual interpretations and contextual social dynamics. This misalignment between LLM's internal decision-making and explicit surface behavior is also consistent with findings observed by prior work \citep{liu2023examining}. Recognizing these complexities is essential for understanding and anticipating the unpredictability in social media engagement.

\begin{table*}[htbp]
\centering
\caption{Agent Reasoning for Content Engagement Analysis}
\small 
\begin{tabular}{lccccp{3.5cm}} 
\toprule
\textbf{Action Type} & \textbf{Total Actions (\%)} & \textbf{Positive (\%)} & \textbf{Neutral (\%)} & \textbf{Negative (\%)} & \textbf{Top 2 Reasoning Categories} \\
\midrule
share\_post     & 1382 (30.1\%) & 91.8 & 6.6 & 1.6  & agreement (46.3\%)\\[-1pt]
                &               &      &     &      & social\_connection (15.7\%) \\
flag\_post      & 1126 (24.6\%) & 13.5 & 15.5 & 71.0 & information\_value (48.8\%)\\[-1pt]
                &               &      &     &      & misinformation (22.4\%) \\
comment         & 880 (19.2\%)  & 96.8 & 2.7 & 0.5  & agreement (29.2\%)\\[-1pt]
                &               &      &     &      & social\_connection (27.6\%) \\
follow\_user    & 719 (15.7\%)  & 98.9 & 0.8 & 0.3  & personal\_interest (27.4\%)\\[-1pt]
                &               &      &     &      & information\_value (24.4\%) \\
like\_post      & 463 (10.1\%)  & 92.0 & 7.8 & 0.2  & social\_connection (33.6\%)\\[-1pt]
                &               &      &     &      & agreement (23.6\%) \\
ignore          & 9 (0.2\%)     & 77.8 & --  & 22.2 & information\_value (36.4\%)\\[-1pt]
                &               &      &     &      & personal\_interest/agreement (18.2\%) \\
unfollow\_user  & 5 (0.1\%)     & 20.0 & 40.0 & 40.0 & agreement (50.0\%)\\[-1pt]
                &               &      &     &      & emotional\_reaction (16.7\%) \\
\bottomrule
\end{tabular}
\label{tab:agent_reasoning_summary}
\end{table*}

\section{Persona Generation Details}
\label{appendix_persona}

Here, we describe the questions that we sampled and generated for the agent users. The generated personas are stored in JSONL format, with each entry containing a unique identifier, a descriptive narrative, and associated behavioral labels.

\subsection{Persona Replication from Human Survey}

The persona generation method begins by transforming structured survey responses collected from Prolific participants into rich, natural language character descriptions suitable for use in agent-based simulations. Each participant’s responses — covering a wide range of personal, demographic, social, and psychological traits—are encoded in JSONL format, where each line corresponds to a different individual. We first put this file into a list of Python dictionaries, each representing a single participant's answers. The preprocessing pipeline then embeds each answer into a templated sentence structure. This includes details such as age, gender, residential background, number of places lived, favorite activities, values, political stance, income, ethnicity, language, education, religion, social tendencies, hobbies, relationship values, personality, future goals, significant life events, friendship values, and hypothetical financial decisions. By expressing these traits in fluent, first-person-style English, the function essentially replicates each participant’s worldview and identity into a lifelike persona that can guide agent behavior in social simulations. In the final step, it iterates through all participant entries, generates the corresponding natural language persona for each one, and writes the enriched data—including both the original responses and the generated description—back into a new JSONL file. This process creates a bridge between raw human survey data and psychologically grounded agent profiles, enabling more realistic and diverse behaviors in multi-agent environments.

\subsection{Synthetic Persona from Agent Bank}

In contrast to the human-annotated personas derived from survey responses, we also generate fully synthetic personas by sampling from a structured question bank, referred to as the Agent Bank. This bank contains a curated set of 23 multiple-choice questions covering key dimensions of identity, background, and social orientation—ranging from age and gender to values, education, hobbies, and political affiliations. Please refer to the code repository for the complete content and answer choices of each of them. Each question is assigned a label and a fixed set of possible answers. To simulate human-like diversity, we construct agent personas by probabilistically sampling answers from these options, sometimes using uniform random choice and other times leveraging carefully constructed distributions to better mirror real-world population dynamics. For instance, age is generated from a normal distribution centered at 35 with bounds clamped between 18 and 60, while gender is sampled from a distribution reflecting approximate societal proportions. In some cases, dependencies between traits are explicitly modeled—for example, primary language is sampled conditionally based on a person’s ethnicity using manually specified probability distributions that reflect linguistic prevalence across ethnic groups. These sampled answers are then assembled into a dictionary of attributes. From this, we use one of two methods to generate natural language persona descriptions. The first method uses a hardcoded template that deterministically weaves the sampled answers into a coherent paragraph, mimicking the style and structure used for real survey-based personas. The second, more dynamic method leverages GPT-4o to produce creative and varied persona descriptions from the same underlying attributes. A carefully crafted system prompt instructs the model to retain every single piece of information from the attribute dictionary while generating a single fluent paragraph in the second person, presenting the result as a believable and detailed backstory. This ensures that each agent maintains a consistent and complete identity while allowing room for stylistic diversity. Ultimately, each synthetic persona is stored as a structured JSON object containing a unique ID, the full natural language description, and the associated label-value pairs, ready to be deployed as agents in downstream simulations.

\subsection{User Generation and Instantiation}

The foundation of the simulation lies in the creation of realistic individual agentic virtual users. Each agent is instantiated with a detailed persona that shapes their online behavior and engagement patterns.

\paragraph{Persona Generation}

As illustrated in Fig.~\ref{fig:fig1}, personas are generated using a combination of predefined questions and sampling from probabilistic distributions stored in an \texttt{agent\_bank} profile collection, inspired by \citet{park2024generative}.
Key demographic attributes such as age, gender, ethnicity, and primary language are assigned probabilistically to mirror real-world distributions. For instance, age follows a normal distribution centered around 35 years, while other attributes are sampled based on predefined probabilities. We disclose these questions and describe more details of the methodology in Appendix~\ref{appendix_persona}. After synthesizing the structured profiles for agents, we construct a natural language description for each of them. This process leverages a mixture of deterministic rules and LLM-based augmentation using \texttt{GPT-4o} \cite{hurst2024gpt} to enhance diversity and realism.



\paragraph{User Instantiation}

Once personas are generated, they are instantiated as agent users within the simulation. Each agent is assigned a unique user ID and a persona profile that includes background details and interest labels. The relational database serves as the backbone for recording agent activities, ensuring persistent storage of interactions, post-engagements, and behavioral updates. This database facilitates dynamic user tracking and enables post-simulation analysis of engagement trends and content spread. We provide more implementation details in Appendix \ref{appendix_database}.

\section{Models And Computational Budget}

We use APIs for proprietary LLM inferences in our experiments. We locally hosted some smaller open-weight LLMs in the early exploratory stages of the projects, on H100 GPUs. We end up consuming approximately \$600 of OpenAI credits, \$300 of Anthropic API, and \$40 of DeepSeek API.

\section{Details of the Human Study and IRB}
\label{app:human_study_details}

\begin{figure*}[t]
    \centering
    \includegraphics[width=\textwidth]{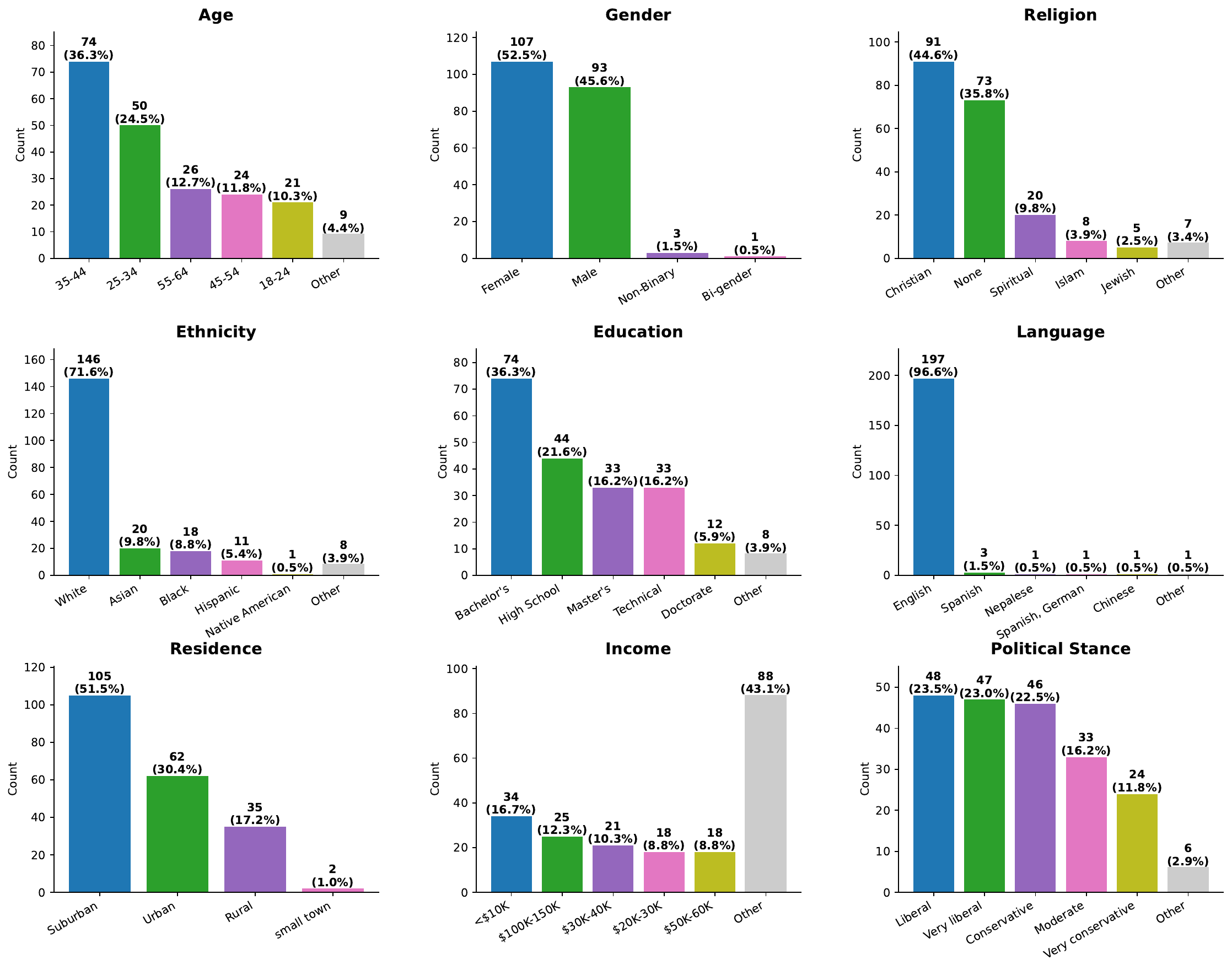}
    \caption{Demographic Distributions of Study Participants.}
    \label{fig:demographic_grid_app}
\end{figure*}

The study was open to 20,240 eligible participants from a larger Prolific population of 232,330, and we collected 204 valid responses from eligible participants. The survey was conducted via Prolific to collect responses from U.S.-based participants fluent in English. Participants were asked to complete a 12-minute survey assessing their demographic characteristics and social media interactions. The survey, hosted on Google Forms, required no software downloads or special device features and was accessible via mobile, tablet, or desktop. Participant recruitment applied custom screening for language, political spectrum, vaccine opinion, and prior participation, ensuring a targeted sample. Responses were collected using Prolific ID via a question at the start of the form, and participants received a completion code upon finishing. Compensation was set at \$2.40 per participant, equivalent to \$12.00/hour, and submissions were manually reviewed before approval. The median completion time was approximately 14.5 minutes. All members on our research team have obtained IRB approval before the human study was conducted. Our study costs a total of \$480 for participant payment and \$160 of the platform fee. Fig.~\ref{fig:demographic_grid_app} shows the complete breakdown of the 9 key demographic distributions of the 204 human participants. This study was performed under approval from the appropriate institutional ethics review board. Full IRB documentation will be made available upon request.

\begin{table*}[htp]
    \centering
    \caption{Demographic groups showing significant differences in engagement patterns between human participants and AI agents (p < 0.05).}
    \label{tab:human_vs_agent}
    \small
    \begin{tabular}{@{\hspace{4pt}}l@{\hspace{8pt}}l@{\hspace{8pt}}l@{\hspace{4pt}}}
        \toprule
        \textbf{Category} & \textbf{Significant Differences} & \textbf{Non-Significant Differences} \\
        \midrule
        Age & 25-34 (shares) & 18-24, 35-44, 45-54, 55-64, 65-74 \\
        Gender & Male (likes, shares) & Female \\
        Religion & Hinduism (likes), Islam (shares) & No Religion, Spiritual, Christianity, Jewish \\
        Ethnic Group & Hispanic/Latino, Black/African (shares), & White/Caucasian, Mixed, Others \\
        & Asian (comments) & \\
        Education & Secondary (shares), Doctorate (likes) & High School, Undergraduate, Technical, Graduate \\
        Income & \$10K-\$20K (comments), \$70K-\$80K (likes) & Various other income brackets \\
        Political Stance & Conservative (shares), Very Conservative (likes) & Very Liberal, Moderate, Liberal, Libertarian \\
        \bottomrule
    \end{tabular}
\end{table*}

\subsection{Per-Demographic Attribute Engagement Pattern}

We also analyzed the reaction patterns between human participants and persona-driven agents grouped by specific demographic attributes such as age, gender, income, ethnicity, etc, as shown in Tab.~\ref{tab:human_vs_agent}.

The analysis of engagement patterns between humans and agents reveals some differences across various demographic groups. Specifically, the age group 25-34 shows notable differences in shares, while males exhibit significant variations in both likes and shares. Among religious groups, Hinduism and Islam display significant differences in likes and shares, respectively. Ethnic groups such as Hispanic or Latino, Black or African American, and Asian show significant differences in shares and comments. Education levels also play a role, with secondary education and doctorate degree holders showing significant differences in shares and likes, respectively. Income levels between \$10,000 - \$19,999 and \$70,000 - \$79,999 show significant differences in comments and likes. Political stances such as Conservative and Very Conservative also exhibit significant differences in shares and likes. In contrast, many other demographic groups, including various age ranges, genders, religions, ethnicities, education levels, income brackets, and political stances, show no significant differences in engagement types. 

\textbf{Overall, out of 52 demographic groups analyzed, 14 show significant differences in one or more engagement types, while 38 do not.} The criteria for significance were based on a p-value of less than 0.05 in statistical comparisons. The results suggest that agents may be more adept at simulating the engagement patterns of "common" or more broadly represented demographic groups in LLM pretraining data, as indicated by the lack of significant differences in many of these groups. We find that out of 52 examined demographic subgroups, only 14 showed statistically significant differences (p < 0.05) in at least one engagement metric (Tab.~\ref{tab:human_vs_agent}). Notable discrepancies appeared in the 25–34 age group (shares) and several religious, ethnic, educational, income, and political categories. However, most demographic groups exhibited no significant differences, suggesting that agents simulate typical engagement behavior more accurately for demographics more prevalent in LLM training data.

\section{An Extended Version of Related Work}

\paragraph{Behavioral Economics and Persuasion Games.}

Our system computationally models a sequential persuasion game with LLM-powered agents conditioned on fine-grained personas \cite{kamenica2011bayesian,gentzkow2017bayesian,acemoglu2023model}. The agents interact within a directed social graph and evolve based on memory and social context. This framework serves as a testbed for studying online behaviors, intervention strategies, and the impact of algorithmic moderation.

\paragraph{AI-Driven Social Simulations and Generative Agents.}

The emergence of large language models (LLMs) has significantly advanced the capabilities of agent-based social simulations, enabling more sophisticated, context-aware interactions. Traditional agent-based modeling relied on predefined rule sets and heuristics, limiting adaptability and realism. Early computational social simulations, such as Schelling's segregation model \cite{schelling1971dynamic}, Sugarscape \cite{epstein1996growing}, and NetLogo-based models \cite{Wilensky1999NetLogo1}, provided insights into social dynamics but lacked the ability to generate nuanced, context-dependent behaviors. 

Recent advances, such as Smallville \cite{park2023generative}, AgentVerse \cite{chen2024agentverse}, Internet-of-Agents \cite{chen2024internet}, and Chirper \cite{chirper}, leverage LLMs to enable generative agents that dynamically respond to evolving contexts. These systems showcase how AI-powered agents can engage in lifelike conversations, form social relationships, and simulate content dissemination patterns. However, despite their ability to generate plausible interactions, generative agents can still exhibit inconsistencies due to biases inherent in LLM training data or limitations in long-term memory and reasoning. By integrating more structured constraints and iterative feedback mechanisms, this work enhances the reliability of agent-based simulations for social science research and policy testing.

\paragraph{Misinformation Spread and Fact-Checking Mechanisms.}

The spread of misinformation on digital platforms has been extensively studied \cite{swire2020public, jerit2020political, wu2019misinformation, islam2020deep}, with empirical evidence showing that falsehoods often propagate more rapidly and broadly than factual information \cite{vosoughi2018spread}. The virality of misinformation is attributed to its emotional appeal, novelty, and the role of engagement-driven algorithms that inadvertently amplify misleading narratives \cite{Pennycook2021ThePO, solovev2022moral}. Addressing this issue has led to the development of multiple fact-checking methodologies, including third-party verification, algorithmic detection, and crowdsourced moderation.

Third-party fact-checking, typically conducted by organizations such as Snopes\footnote{\url{https://www.snopes.com/}}, PolitiFact\footnote{\url{https://www.politifact.com/}}, or Google's partnerships with external organizations \cite{googleIndia, googleAus}, provides authoritative assessments but faces challenges in scalability and timeliness \cite{zannettou2019web, uscinski2013epistemology, marietta2015fact}. Crowdsourced fact-checking, such as X's Community Notes,\footnote{\url{https://communitynotes.x.com/guide/en/}} on the other hand, leverages collective intelligence \cite{panizza2023online} but introduces risks related to expertise and susceptibility to group biases \cite{saeed2022crowdsourced, pennycook2021shifting}. There is no consensus on which fact-checking approach is more effective, nor is it well-understood how different moderation strategies interact. This study addresses this gap by leveraging LLM-driven simulations to evaluate different fact-checking mechanisms within controlled environments. By testing various moderation strategies in a scalable, repeatable manner, this work provides insights into the comparative efficacy of community-based, third-party, and hybrid fact-checking interventions in mitigating misinformation.

\paragraph{Simulations as Tools for Policy and Platform Governance}

The use of computational simulations as decision-support tools has been well-established in domains such as epidemiology \cite{currie2020simulation, lorig2021agent}, economics \cite{axtell2022agent}, and public policy \cite{qu2024performance}. By enabling scenario testing before real-world implementation, simulations help policymakers anticipate the consequences of interventions \cite{charalabidis2011enhancing}. In the context of social media governance, AI-driven simulations present an emerging opportunity to evaluate moderation strategies, optimize intervention policies, and test the societal impact of algorithmic changes before deployment.

Recent discourse around AI governance emphasizes the need for proactive measures to ensure platform accountability and transparency \cite{landau2024challenging}. Regulatory bodies and platform operators are increasingly exploring ways to assess the impact of interventions such as content moderation adjustments, ranking algorithm changes, and misinformation mitigation strategies before rolling them out at scale. To this end, our research introduces AI-driven social simulations as a novel framework for governance experimentation. By simulating diverse social environments and misinformation dynamics, we provide an approach that offers a scalable, controlled setting for testing policy interventions. This methodology aligns with the growing call for algorithmic auditing and regulatory sandboxes, providing a novel tool for both researchers and policymakers to refine governance strategies before real-world application.

\section{Database Schema}
\label{appendix_database}

In this section, we describe the database schema that we developed to store and keep track of all the data generated by each simulation run.

This relational SQL database schema is designed to support a social media simulation in which LLM-powered AI agents mimic user behaviors. The database captures and organizes user-generated content, interactions, and system-level processes in detail. The \texttt{users} table stores individual user profiles, including metadata such as personas, background labels, influence scores, and engagement metrics. Posts authored by users are managed in the \texttt{posts} table, which records content details, interaction counts (likes, shares, flags, comments), and moderation or fact-check statuses. Social relationships are modeled through the \texttt{follows} table, which tracks follower-followed connections. User engagement actions, such as creating content or reacting to posts, are logged in the \texttt{user\_actions} table. Comments on posts are separately recorded in the \texttt{comments} table with their associated metadata. Community moderation is facilitated via the \texttt{community\_notes} and \texttt{note\_ratings} tables, enabling users to contribute interpretive notes and rate their helpfulness. System moderation decisions are logged in \texttt{moderation\_logs}. The \texttt{fact\_checks} table provides detailed verdicts and rationales from fact-checking processes. To simulate memory and reasoning for AI agents, \texttt{agent\_memories} track the content and importance of internal memories, with timestamps and decay factors. The \texttt{spread\_metrics} table quantifies the virality and diffusion dynamics of each post over time steps, including derived interaction statistics and takedown decisions. Exposure to content is tracked at the user level in the \texttt{feed\_exposures} table, supporting the analysis of information visibility and reach. Together, these schemas capture a detailed and interconnected view of simulated social media dynamics, grounded in observable user behavior and system responses.

\section{Agent Action Space}
\label{appendix_action_space}

\begin{figure}[htp]
    \centering
    \includegraphics[width=0.5\textwidth]{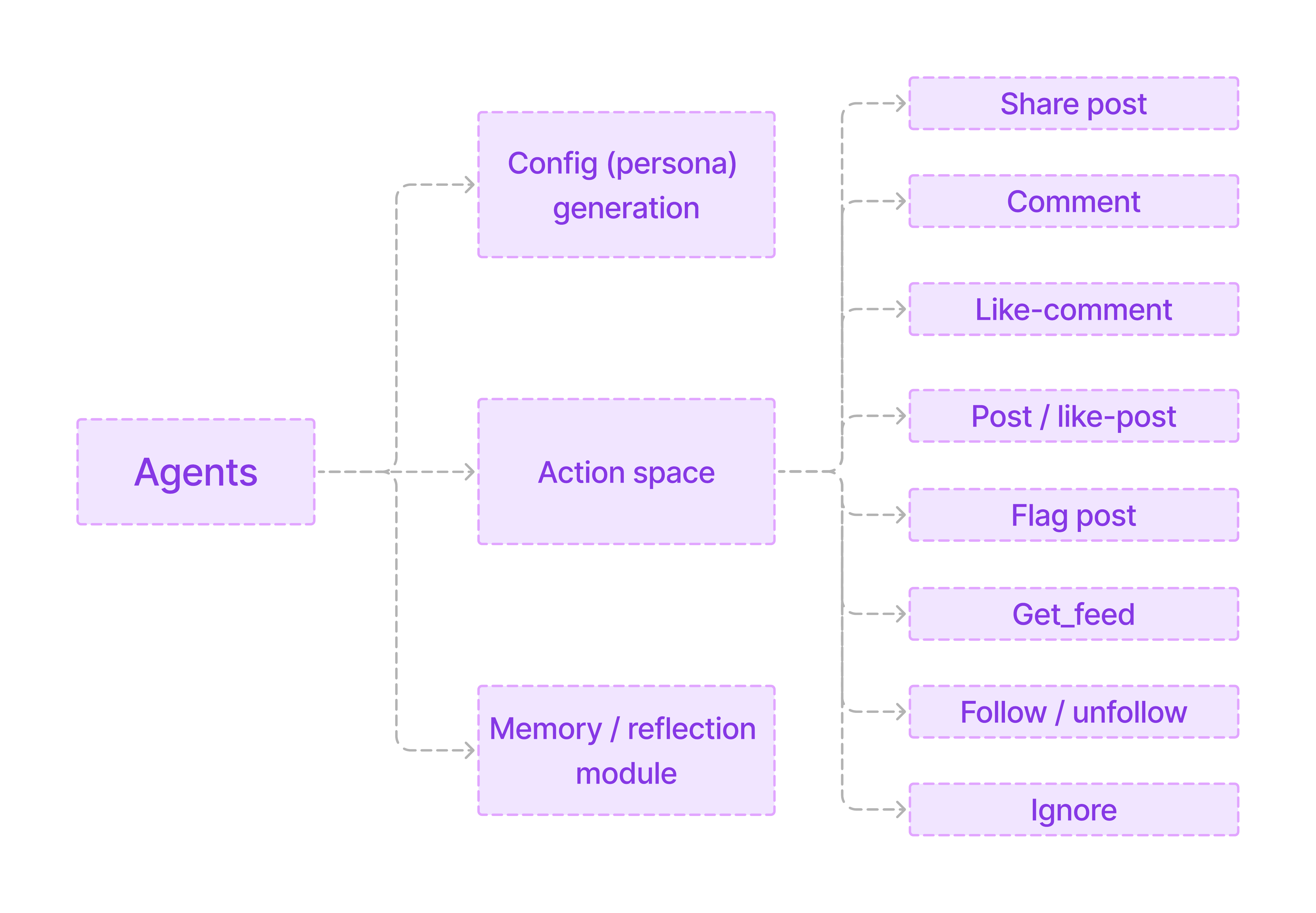}
    \caption{Agent's action space.}
    \label{fig:action_space}
\end{figure}

In our simulated social media environment, each agent—powered by a large language model—is instantiated with a predefined action space that governs its interactions within the platform. These agents are configured with unique personas via a configuration (persona) generation module and endowed with a memory/reflection module that allows them to recall and adapt based on past experiences. The Action space outlines the full spectrum of behaviors an agent can exhibit: they can share posts, comment or like comments, create or like posts, and flag inappropriate content. Additionally, agents can retrieve their feeds, follow or unfollow other users, or ignore content or interactions altogether. These discrete actions simulate realistic user behavior and social dynamics, enabling rich, emergent interactions in the environment.

\subsection{Agent Decision-Making Process}

The agent decision-making process is governed by structured interactions between feed presentation, memory recall, and reasoning mechanisms.

\paragraph{Feed Presentation}
Each agent’s feed aggregates posts from followed users, supplemented by additional trending content and news articles. On average, one in ten posts is from the NewsGuard dataset and contains misinformation. Posts are displayed with metadata such as engagement counts (likes, comments, shares) and fact-checking signals (flags, community notes, third-party verdicts). This metadata provides context for the agent’s engagement decisions.

\paragraph{Memory and Reflection Module}

We implement an \texttt{AgentMemory} module which manages the memory and reflection capabilities of each agent. Memories are categorized into \textit{interactions} (e.g., past engagements) and \textit{reflections} (high-level insights derived from past behaviors). Each piece of memory is assigned an importance score, which decays over time unless reinforced by further interactions. The decay function ensures that long-term behaviors emerge naturally based on experience. Please refer to Appendix~\ref{app:memory} for more details of the Memory module.

Periodically, agents generate reflections based on recent interactions. These reflections help detect behavioral patterns, relationship dynamics, and potential biases, influencing future content engagement and decision-making.

\paragraph{Agent Decision-Making and Action Execution}

Agents make decisions based on a combination of persona-driven heuristics, memory retrieval, and reasoning prompts. The \texttt{AgentPrompts} module formulates structured decision prompts, guiding agents through content engagement options such as liking, sharing, or flagging a post. When engaging, agents provide reasoning for their actions, influenced by (1) \textbf{Personal Beliefs and Persona Traits}: Agents weigh content credibility based on their ideological stance and historical preferences, (2) \textbf{Engagement Signals}: Highly engaged posts are more likely to be reshared due to social validation effects, and (3) \textbf{Fact-Checking Feedback}: Agents integrate fact-checking signals into their reasoning, adjusting their trust in flagged content accordingly.

Once a decision is made, the agent's action gets recorded in the relational database, along with updated post metrics, engagement, and new memories. The importance of each interaction is evaluated based on emotional intensity, action strength, and alignment with the agent’s goals.

\section{Detailed Experiment Configuration}
\label{appendix_exp_config}

We describe our experimental settings and configurable variables in more detail in this section. In our simulation, we model a dynamic social network of an arbitrary number of (practically in our experiments, up to over 200) LLM-driven agents interacting over the course of a number of discrete time steps. The simulation loop follows a structured core cycle that includes initializing the environment, assigning new users probabilistically (though this one could be disabled in certain runs), content creation, feed-based reactions, and periodic reflective updates. Each agent is instantiated from detailed persona descriptions provided via an external JSONL file, and operates using the GPT-4o engine with a decoding temperature of 1.0 to promote diversity in generated responses. Agents can be configured to create original posts independently, or they can be prompted to only respond to posts, depending on the setting. Once the simulation environment is initiated, an agent's feed consists of a mixture of up to a default of 15 posts from followed users and 10 from non-followed users, drawn from a pool that includes up to 20 injected news items per run. Initial social ties are sparse, with a 10\% probability of following another user at initialization, and new user addition and follow behaviors are disabled during the simulation. All of the above numbers are configurable. The experiment evaluates one of the four fact-checking intervention modes described in Section~\ref{sec3:content-moderation}, combining both third-party and community-based mechanisms. For each step, if a fact-checking agent is enabled, then a number of posts are selected for potential moderation, with fact-checking outputs generated using a low-temperature (0.3) setting and required to include reasoning. Thresholds are specified for flagging and note-taking behavior if moderation is set to active. Periodically, agents reflect on their recent interactions, update memory states, and check their internal objectives, offering a framework for studying emergent behavior, information diffusion, and intervention efficacy in artificial societies.

\section{Details of the Memory Module}
\label{app:memory}

Memory relevance is computed as:
\[
\text{Relevance} = \text{Importance} \times \text{Decay}
\]

The decay factor is defined as:
\[
\text{Decay} = \max(0, \text{PrevDecay} - \alpha \Delta t)
\]
where $\alpha$ is the decay rate (default 0.1), and $\Delta t$ is the time (in days) since last access. New memories start with $\text{PrevDecay} = 1.0$.

A memory is considered relevant if $\text{Relevance} \geq 0.3$. Both Importance and Decay are in $[0,1]$.

\paragraph{Importance Scoring.} Each memory has a base importance score of 0.5. This is increased by 0.1 for each keyword match (up to a max of 1.0) from the following semantic categories:

\begin{itemize}
    \item Emotional: \textit{love, hate, angry, happy, sad}
    \item Action: \textit{achieved, failed, learned, discovered}
    \item Relationship: \textit{friend, follow, connect, share}
    \item Goal: \textit{objective, target, aim, purpose}
\end{itemize}

Let $k$ be the number of keyword matches in the memory content. Then:
\[
\text{Importance} = \min(1.0, 0.5 + 0.1k)
\]

This value is combined with the decay factor to compute final relevance.


\begin{figure}[htp]
    \centering
    \includegraphics[width=0.45\textwidth]{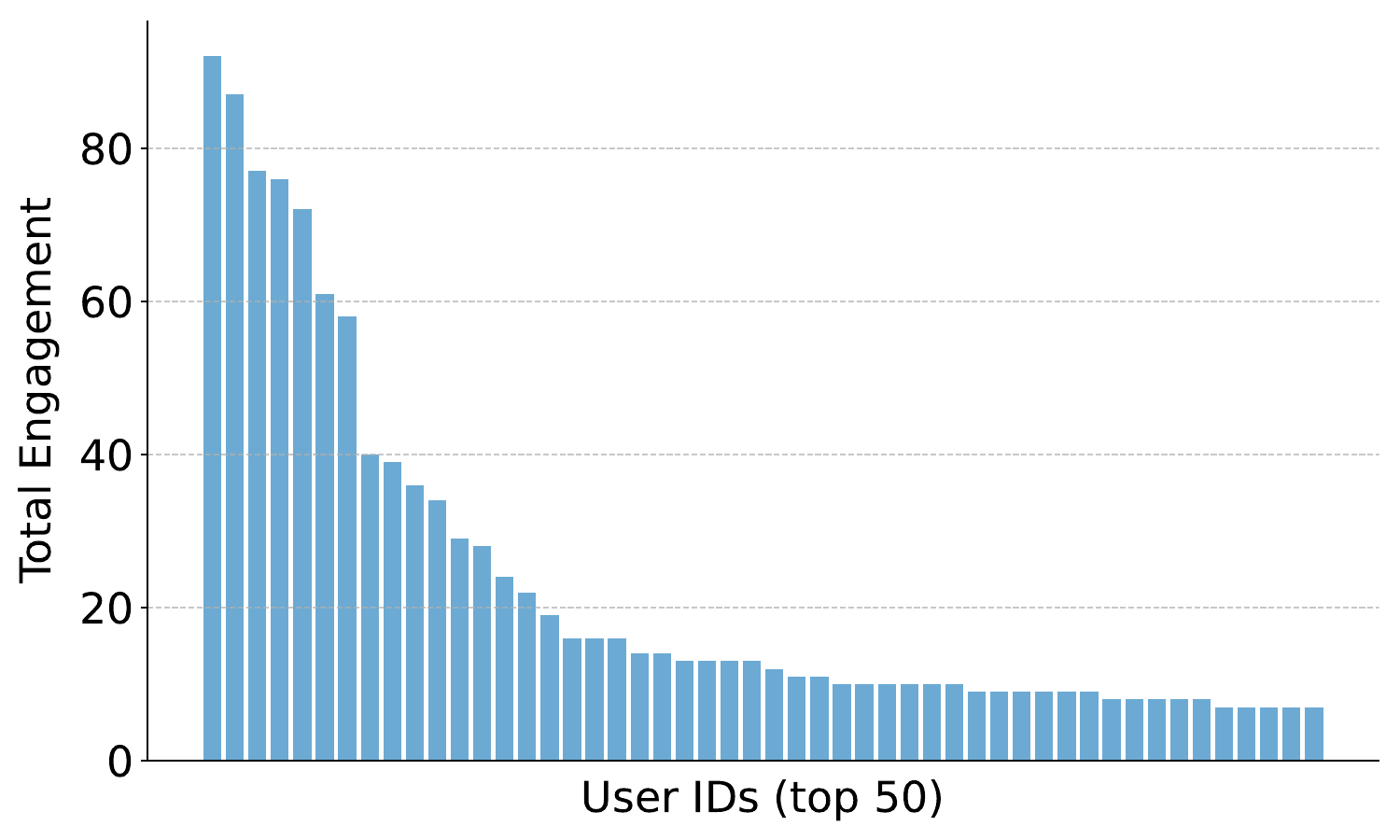}
    \caption{Top 50 users with the highest engagement.}
    \label{fig:top50_users}
\end{figure}

\section{Total Engagement Comparison}
\label{app:engagement_analysis_detailed}

\begin{figure*}[t]
    \centering
    \includegraphics[width=\textwidth]{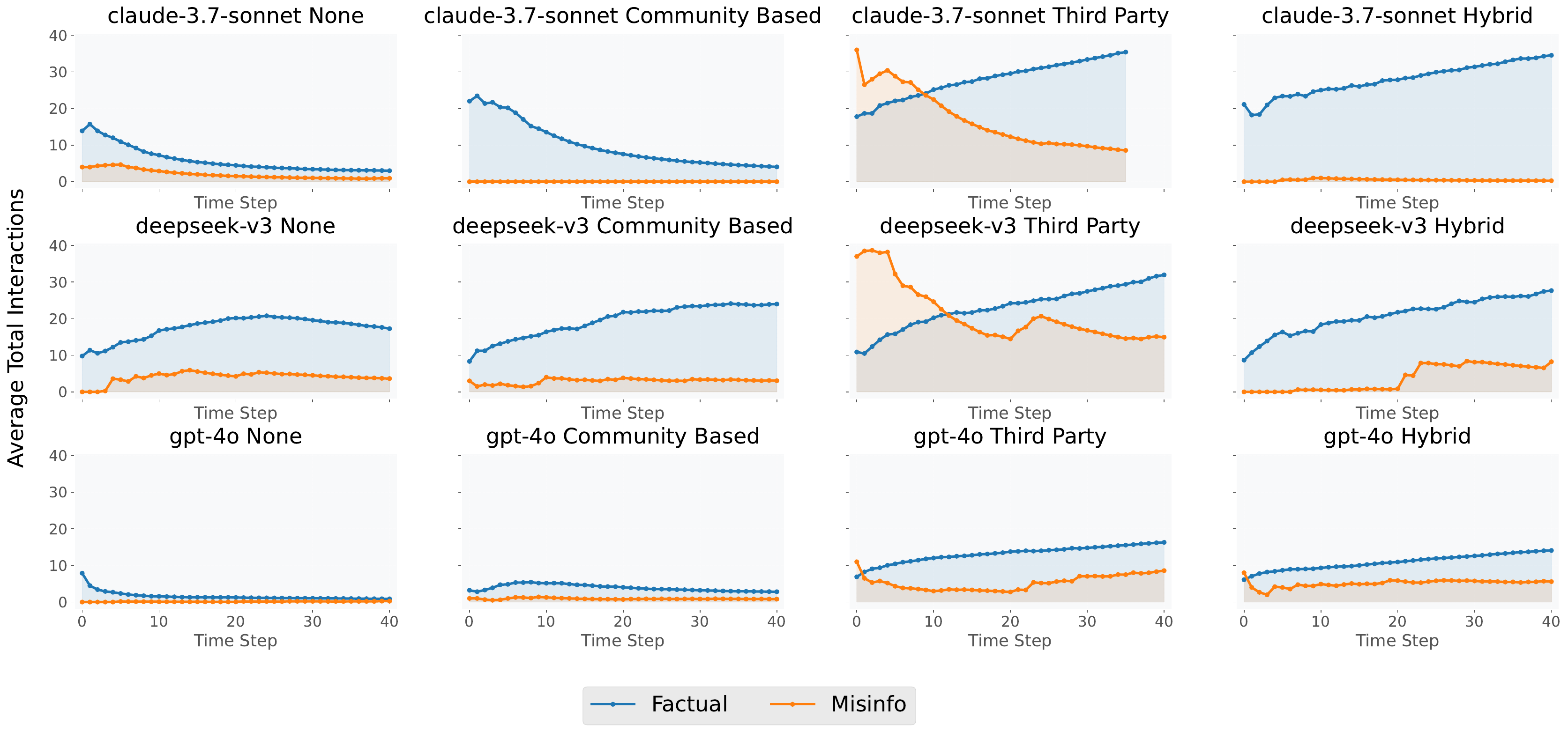}
    \caption{Total Engagement Across All Models and Fact Check Modes}
    \label{fig:all_engagement_comparison_total}
\end{figure*}

\begin{figure*}[ht]
    \centering
    \includegraphics[width=\textwidth]{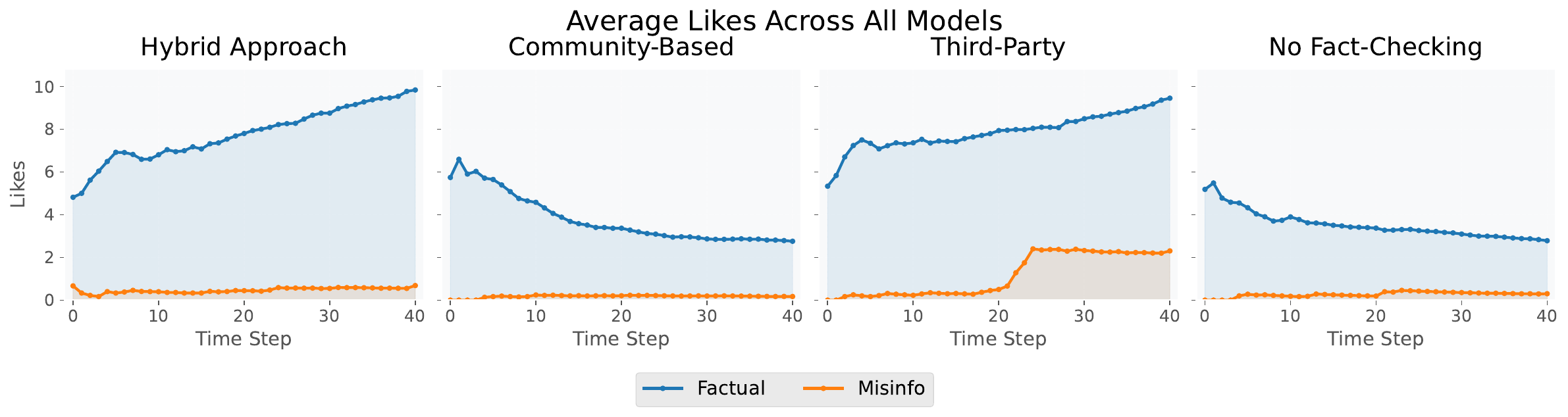}
    \caption{Average Number of Likes}
    \label{fig:average_num_likes}
\end{figure*}

Fig.~\ref{fig:all_engagement_comparison_total} tracks average interactions with factual and misinformative posts across time steps 0-40 for three LLMs: Claude-3.7-Sonnet, DeepSeek-V3, and GPT-4o over four fact-checking types as defined in the experiment configurations: None, Community Based, Third Party, and Hybrid. 
Each subplot shows how user engagement with true and false content evolves. 

\subsection{Total Interactions: Claude-3.7-Sonnet}
\textbf{"None", "Community Based"}: There is little to no interaction with false news. Interaction with factual posts decreases over time. 
\newline \textbf{"Third Party"}: Initially, there is higher interaction with false news than factual content. This reverses as interaction with misinformation decreases and with factual content increases.
\newline \textbf{"Hybrid"}: There is little to no interaction with false news. There is a steady increase in interaction with factual content. 
\subsection{Total Interactions: DeepSeek-V3}
\textbf{"None", "Community Based", "Hybrid"}: There is little interaction with misinformation. Interaction with factual content starts off strongly and increases over time. 
\newline \textbf{"Third Party"}: Initially, there is much higher interaction with false news than factual content. Over time, interaction with misinformation decreases and with factual content increases, but interaction with misinformation is still close to factual content interaction. 
\subsection{Total Interactions: GPT-4o}
\textbf{"None", "Community Based"}: There is little interaction with both factual and false news. 
\newline \textbf{"Third Party", "Hybrid"}: There is some interaction with false news, though less than the consistent and increasing interaction with factual content. 

\begin{figure}[ht]
    \centering
    \includegraphics[width=0.5\textwidth]{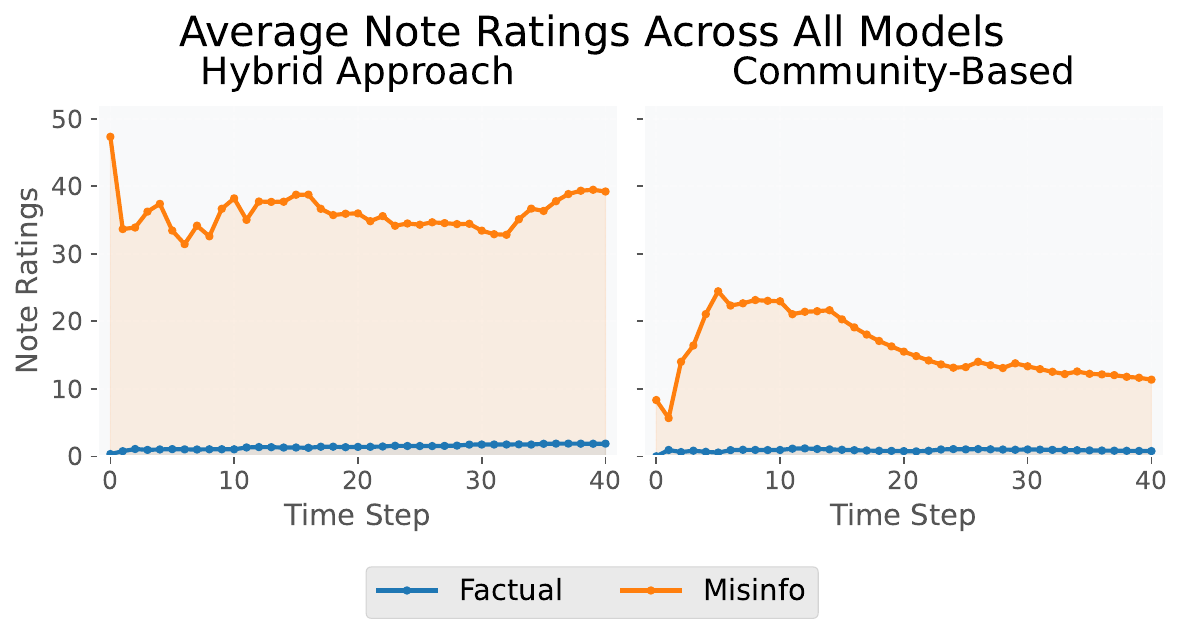}
    \caption{Average Number of Note Ratings}
    \label{fig:average_num_note_ratings}
\end{figure}

\subsection{Cross Model Comparisons}
\textbf{"None"}: All models have very little interaction with false news.  DeepSeek is the only model that increases interaction with factual content, while the others decrease over time. 
\newline \textbf{"Community Based"}: All models continue to have very little interaction with false news. Again, DeepSeek is the only model that increases interaction with factual content over time. 
\newline \textbf{"Third Party"}: All experiments initially start with a higher level of interaction with misinformation than interaction with factual information. This trend reverses across all models over time. However, the level of difference in interaction with misinformation and factual information is most pronounced in Claude. Other models have less differentiation. 
\newline \textbf{"Hybrid"}: Claude has the lowest levels of interaction with false news as well as the highest and most consistent levels of interaction with factual content. DeepSeek initially displays the same trend, while experiencing increasing misinformation engagement in later time steps. GPT-4o has constant interaction with different content types, though slightly higher with factual content. 

\subsection{Potential Claims}
\textbf{Claude-3.7-Sonnet}: Under the "None" and "Community Based" settings, Claude shows almost no interaction with misinformation. However, interactions with factual content decline steadily over time. This suggests that in the absence of strong moderation signals, even factual content loses traction and user engagement drops.

With "Third Party" moderation, there is initially more engagement with misinformation than with factual content. Over time, this trend reverses: misinformation engagement declines while factual engagement increases. This shift implies that authoritative third-party intervention can realign attention toward accurate information.

Under the "Hybrid" setting, Claude exhibits the strongest performance. Misinformation is almost completely suppressed, while factual content steadily gains engagement throughout the simulation. This indicates that Claude, when acting as an agent, is highly responsive to layered, multi-source moderation and can maintain a sustained pro-factual trajectory when given comprehensive oversight.

\textbf{DeepSeek-V3}: DeepSeek behaves differently from Claude. In the "None," "Community Based," and "Hybrid" settings, it maintains low misinformation engagement while steadily increasing interactions with factual content. This pattern suggests DeepSeek may have a stronger default tendency toward promoting truthful content, even with minimal intervention.

Under the "Third Party" condition, however, the simulation starts with high misinformation engagement. Although this decreases over time and begins to converge with factual engagement, misinformation remains close in magnitude. Unlike Claude, DeepSeek does not show a strong corrective response to third-party fact-checking, suggesting lower sensitivity to external moderation.

\textbf{GPT-4o}: GPT shows low overall engagement with both factual and false news under the "None" and "Community Based" regimes. This could reflect a more cautious content-sharing dynamic when no clear verification cues are available.

Under both the "Third Party" and "Hybrid" settings, factual engagement consistently increases, while misinformation remains relatively low. Though the effect is less dramatic than with Claude, GPT demonstrates a stable alignment with accurate content in the presence of reliable fact-checking mechanisms, suggesting moderate responsiveness to external moderation.

Our findings may suggest that layered, hybrid fact-checking is most effective overall, especially for models that are more responsive to external moderation like Our findings may suggest that layered, hybrid fact-checking is most effective overall, especially for models that are more responsive to external moderation like Claude and GPT. In contrast, models like DeepSeek may require tailored or persistent strategies to achieve similar alignment outcomes. and GPT. In contrast, models like DeepSeek may require tailored or persistent strategies to achieve similar alignment outcomes.

\begin{figure}[ht]
    \centering
    \includegraphics[width=0.5\textwidth]{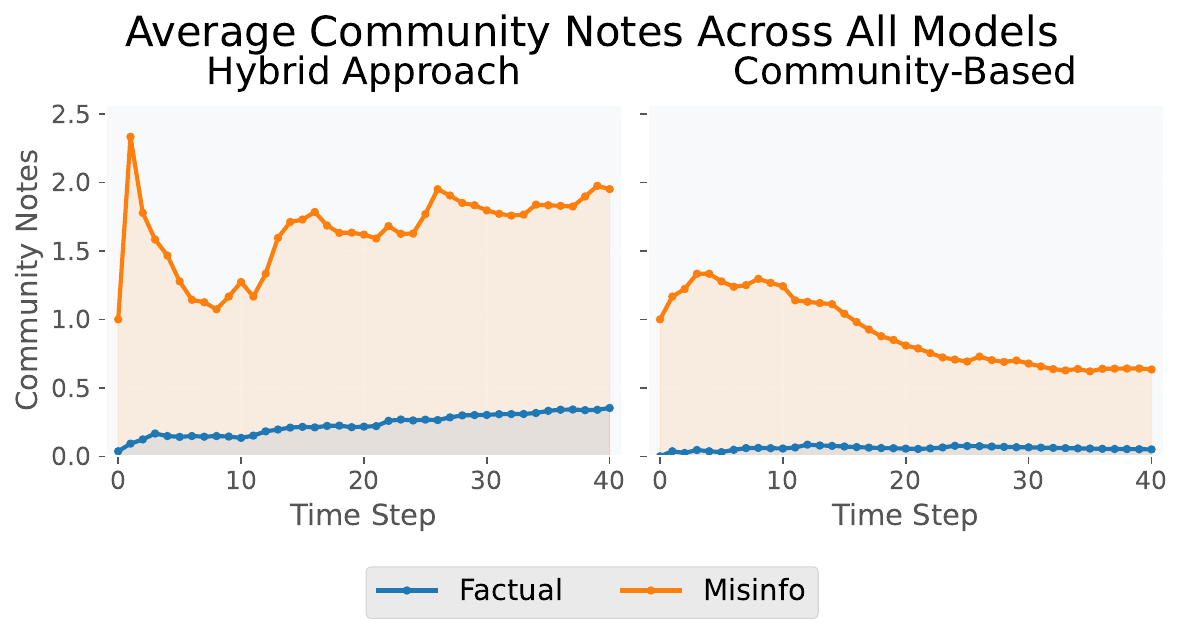}
    \caption{Average Number of Notes}
    \label{fig:average_num_notes}
\end{figure}

\section{Disaggregated Engagement Comparison}
We've decided to further explore interactions of all models across misinformative and factual content through five allowed actions: comments, likes, note ratings, community notes, and shares. 

\subsection{Average Comments}
Fig.~\ref{fig:average_num_comments} shows Average Comments across all models. In "Hybrid", comments on factual content steadily increase over time, while misinformation receives minimal and flat engagement. In "Community Based" and "None", misinformation remains low and nearly flat, while factual content is slightly higher. In "Third-Party", there is a steady increase in comments on factual content, while comments on false news start high and decrease. Strong moderation (like hybrid) seems to promote public discussion of factual content while limiting commentary on misinformation. 

\begin{figure*}[t]
    \centering
    \includegraphics[width=\textwidth]{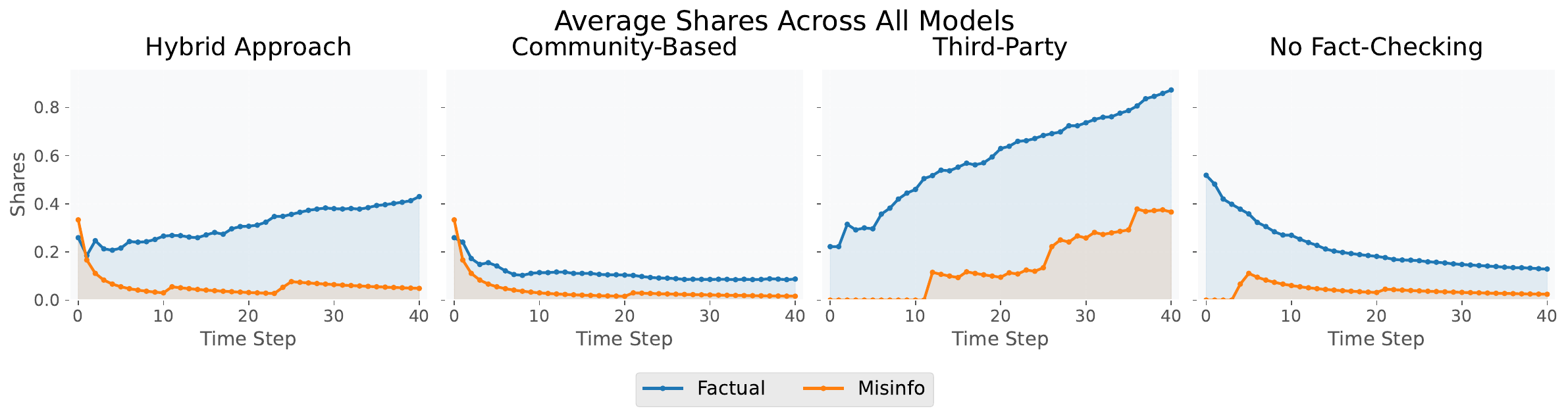}
    \caption{Average Number of Shares}
    \label{fig:average_num_shares}
\end{figure*}

\begin{figure*}[t]
    \centering
    \includegraphics[width=\textwidth]{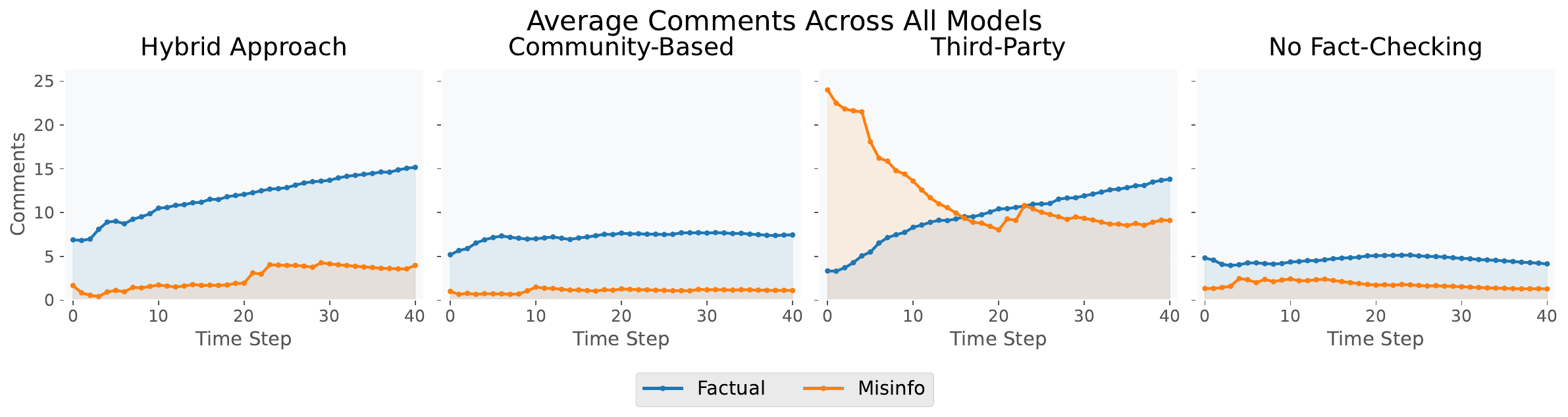}
    \caption{Average Number of Comments}
    \label{fig:average_num_comments}
\end{figure*}

\subsection{Average Likes}
Fig.~\ref{fig:average_num_likes} shows Average Likes across all models. Likes across all fact-checking modes have high engagement over time for factual information, with minimal engagement with misinformation. For "Hybrid" and "Third Party" approaches, likes trend up, while for "Community-Based" and "No Fact Checking", likes trend down over time. Interestingly, while there is almost no engagement in other modes, there remains some likes for misinformation in "Third-Party". Hybrid moderation seems to foster the strongest positive sentiment toward factual content, with the lowest disengagement with false news.

\subsection{Average Note Ratings}
Fig.~\ref{fig:average_num_note_ratings} shows the average Note Rating count across all models. 
There is a higher amount of note ratings in "Hybrid" than "Community-Based". Additionally, expected behavior is that there are few notes to rate for factual information. 

\subsection{Average Community Notes}
Fig.~\ref{fig:average_num_notes} shows the average Community Note count across all models. In "Hybrid", there is a steady growth of community notes for misinformation for the majority of the time steps. In "Community-Based", there is actually a steady decline. There are also no notes for factual information in "Community-Based", while there are a few but steadily growing notes in "Hybrid". 

Note Ratings and Community Notes suggest that a "Hybrid" approach best stimulates robust feedback loops for users, including interaction with all types of information and interaction with other users' notes. 

\subsection{Average Shares}
Fig.~\ref{fig:average_num_shares} shows the final action possible for agents, Average Share count across all models. The number of average shares is generally low, being less than 1 per post. In this proportion, for "Hybrid", "Community-Based", and "None" approaches, there is a fast increase to decrease and constant low in the sharing of misinformation. "Hybrid" has gradually increasing shares for factual information, while "Community-Based" and "None" slowly decrease, though "None" has more overall shares. "Third-Party" experiences the most overall shares, and the sharpest increase in shares of factual information over time as well. Overall, strong fact-checking encourages users to amplify factual content, whereas misinformation rarely achieves viral spread across any setting.

\section{Emotional Salience of False vs. True News}
\label{app:emotional_salience}

To validate that our dataset reflects the known emotional drivers of rumor propagation, we conduct an auxiliary analysis comparing the emotional salience of real versus false news. Using GPT-4o as an annotator, we evaluated articles along established psychological dimensions of viral misinformation. We sampled 1,353 real news articles and paired them with 1,353 false articles from the NewsGuard dataset for a balanced 1:1 comparison.

Tab~\ref{tab:emo-scores} reports average scores across six dimensions. False news consistently exhibits higher emotional intensity, fear, anger, and urgency, while being perceived as less credible.  

\begin{table*}[ht]
\centering
\begin{tabular}{lcccc}
\hline
Dimension & Real & False & Difference & Effect Size \\
\hline
Emotional Intensity \cite{berger2012makes} & 5.0 & 7.7 & +2.7 & +55\% \\
Fear/Anxiety \cite{vosoughi2018spread} & 3.6 & 7.8 & +4.2 & +116\% \\
Anger/Outrage \cite{chuai2020anger} & 3.1 & 7.6 & +4.5 & +145\% \\
Urgency \cite{song2023message} & 4.3 & 6.9 & +2.6 & +60\% \\
Credibility \cite{mang2024source} & 6.9 & 3.0 & -3.8 & -55\% \\
Overall Rumor Potential \cite{zhang2021conspiracy} & 4.0 & 8.5 & +4.5 & +110\% \\
\hline
\end{tabular}
\caption{Comparative emotional profile of real vs. false news articles (1–10 scale).}
\label{tab:emo-scores}
\end{table*}

False news exhibits consistently higher levels of fear, anger, urgency, and emotional intensity, while being rated substantially lower in credibility. To complement this, Tab.~\ref{tab:high-emotion} summarizes the proportion of articles that scored above 7/10 on each dimension. False news dominates across all measures, with particularly strong differences in anger and fear. 

\begin{table*}[ht]
\centering
\begin{tabular}{lccc}
\hline
Metric & Real News & False News & Ratio \\
\hline
High Emotional Intensity ($\geq$7/10) & 335 (24.8\%) & 1,293 (95.6\%) & 3.9x \\
High Fear/Anxiety & 236 (17.4\%) & 1,157 (85.5\%) & 4.9x \\
High Anger/Outrage & 186 (13.7\%) & 1,170 (86.5\%) & 6.3x \\
High Rumor Potential & 175 (12.9\%) & 1,310 (96.8\%) & 7.5x \\
\hline
\end{tabular}
\caption{Distribution of high-emotion articles across real and false news.}
\label{tab:high-emotion}
\end{table*}

These findings confirm that the NewsGuard dataset is well-calibrated to reflect the emotional salience typical of real-world misinformation. Specifically, false news demonstrates 7.5x higher viral potential, 145\% greater anger/outrage, and 116\% greater fear/anxiety than real news, consistent with prior work on rumor spread dynamics. This provides additional support for the realism of our simulation experiments.

\section{Prompts}
\label{add:prompts}

\subsection{Post Creation Prompt}


\begin{lstlisting}[
  basicstyle=\ttfamily\footnotesize,
  breaklines=true,
  breakatwhitespace=false,
  columns=flexible,
  keepspaces=true,
  backgroundcolor=\color{blue!5!white},
  frame=none,
  xleftmargin=0pt
]

"""
Given:
 - persona: str,
 - memories_text: str,
 - recent_posts_text: str,
 - feed_text: str
"""

Create a social media tweet for a user with the following characteristics:

Background: {persona}

Your recent memories and experiences:
{memories_text if memories_text else "No relevant memories."}

Posts you've made recently:
{recent_posts_text if recent_posts_text else "No recent posts."}

Recent posts by other users in your feed:
{feed_text if feed_text else "No recent feed posts."}

The post should be authentic to the user's persona and background and can reference your past experiences. Keep it concise and suitable for a social media platform.

IMPORTANT: 
- Avoid repeating similar topics or themes from your recent posts
- Try to bring fresh perspectives or discuss different aspects of your interests
- Feel free to engage with or reference one or more recent posts from your feed when relevant
- If there's breaking news in your feed, consider engaging with it if it aligns with your interests, whether you agree or disagree

You don't need to always use emojis every time you write something.

Consider the posts that you have made recently. 
Try to diversify your content and style. For example, avoid always starting a post with the same phrase like "just ..."

The post you are about to create is:
\end{lstlisting}

\subsection{Feed Reaction Prompt}

\begin{lstlisting}[
  basicstyle=\ttfamily\footnotesize,
  breaklines=true,
  breakatwhitespace=false,
  columns=flexible,
  keepspaces=true,
  backgroundcolor=\color{blue!5!white},
  frame=none,
  xleftmargin=0pt
]

def create_feed_reaction_prompt(
        persona: str,
        memories_text: str,
        feed_content: str,
        reflections_text: str = "",
        experiment_type: str = "third_party_fact_checking",
        include_reasoning: bool = False
    ) -> str:
        # Base prompt that's common across all experiment types
        base_prompt = f"""You are browsing your social media feed as a user with this background:
{persona}

Recent memories and interactions:
{memories_text if memories_text else "No relevant memories."}

Your feed:
--------------------------------
{feed_content if feed_content else "No recent feed posts."}
--------------------------------

Your past reflections:
{reflections_text if reflections_text else "N/A"}

Based on your persona, memories, and the content you see, choose how to interact with the feed.
"""
        if not experiment_type:
            raise ValueError("Experiment type is required")

        # Add experiment-specific instructions and valid actions
        if experiment_type == "no_fact_checking":
            base_prompt += """
Valid actions:
- like-post // [post_id]
- share-post // [post_id]
- comment-post // [post_id] with [content], limited to 250 characters
- ignore

Interact with posts and users based on your interests and beliefs. 
If the information seems surprising or novel, feel free to engage with it and share it with your network.
"""
        elif experiment_type == "third_party_fact_checking":
            base_prompt += """
Valid actions:
- like-post // [post_id]
- share-post // [post_id]
- comment-post // [post_id] with [content], limited to 250 characters
- ignore
"""
        elif experiment_type == "community_fact_checking":
            base_prompt += """
You can add community notes to posts that you think need additional context or fact-checking.
You can also rate existing community notes as helpful or not helpful based on their accuracy and usefulness.

Valid actions:
- like-post // [post_id]
- share-post // [post_id]
- comment-post // [post_id] with [content], limited to 250 characters
- add-note // [post_id] with [content] - Add a community note to provide context or fact-checking
- rate-note // [note_id] as [helpful/not-helpful] - Rate existing community notes
- ignore

If you see existing community notes on a post, first consider rating them as helpful or not helpful, and then add your own note ONLY if you have additional context to provide.
"""
        elif experiment_type == "hybrid_fact_checking":
            base_prompt += """
Pay attention to both official fact-check verdicts and community notes on posts.
You can add your own community notes and rate existing ones, while also considering official fact-checks.

Valid actions:
- like-post // [post_id]
- share-post // [post_id]
- comment-post // [post_id] with [content], limited to 250 characters
- add-note [post_id] with [content] - Add a community note to provide context or fact-checking
- rate-note [note_id] as [helpful/not-helpful] - Rate existing community notes
- ignore
"""

        base_prompt += """
THESE ARE THE ONLY VALID ACTIONS YOU CAN CHOOSE FROM.
"""

        # Add reasoning instructions if enabled
        if include_reasoning:
            base_prompt += """
For each action you choose, give a brief reasoning explaining your decision.
"""

        base_prompt += """
Respond with a JSON object containing a list of actions. For each action, include:
- action: The action type from the valid actions list
- target: The ID of the post/user/comment/note (not needed for 'ignore')
- content: Required for comment-post and add-note actions
"""

        # Add reasoning field to example if enabled
        if include_reasoning:
            base_prompt += """
- reasoning: A brief explanation of why you took this action
"""

        # Add note_rating field for relevant experiment types
        if experiment_type in ["community_fact_checking", "hybrid_fact_checking"]:
            base_prompt += """
- note_rating: Required for rate-note actions ("helpful" or "not-helpful")
"""

        # Example response
        if include_reasoning:
            base_prompt += """
Example response:
{
    "actions": [
        {
            "action": "like-post",
            "target": "post-123",
            "reasoning": "This post contains valuable information"
        },
        {
            "action": "share-post",
            "target": "post-123",
            "reasoning": "I want to spread this important news"
        }
    ]
}"""
        else:
            base_prompt += """
Example response:
{
    "actions": [
        {
            "action": "like-post",
            "target": "post-123"
        },
        {
            "action": "share-post",
            "target": "post-123"
        }
    ]
}"""

        return base_prompt

\end{lstlisting}

\subsection{Reflection Prompt}

\begin{lstlisting}[
  basicstyle=\ttfamily\footnotesize,
  breaklines=true,
  breakatwhitespace=false,
  columns=flexible,
  keepspaces=true,
  backgroundcolor=\color{blue!5!white},
  frame=none,
  xleftmargin=0pt
]

Based on your recent experiences as a social media user with:

Background: {persona}

Recent memories and experiences:
{memory_text}

Reflect on these experiences and generate insights about:
1. Patterns in your interactions
2. Changes in your relationships
3. Evolution of your interests
4. Potential biases or preferences you've developed
5. Goals or objectives you might want to pursue

Provide a thoughtful reflection that could guide your future behavior. Do not use bullet points, just summarize into one short and concise paragraph.

\end{lstlisting}

\subsection{Fact-checking Prompt}

\begin{lstlisting}[
  basicstyle=\ttfamily\footnotesize,
  breaklines=true,
  breakatwhitespace=false,
  columns=flexible,
  keepspaces=true,
  backgroundcolor=\color{blue!5!white},
  frame=none,
  xleftmargin=0pt
]

"""
Given:
 - post_content: str,
 - community_notes: str,
 - engagement_metrics: dict
"""

Please fact-check the following social media post:

Content: {post_content}

Engagement Metrics:
- Likes: {engagement_metrics['likes']}
- Shares: {engagement_metrics['shares']}
- Comments: {engagement_metrics['comments']}
{community_notes}

Please analyze this content and provide:
1. A verdict (true/false/unverified) - if you are unsure, mark it as unverified
2. A detailed explanation of your findings
3. Your confidence level (0.0 to 1.0)
4. List of sources consulted

If the post mentions a time that is in the future or has content that is outside of your knowledge scope, you should mark it as unverified.
For obvious misinformation, you should mark it as false.

Format your response as a structured verdict with these components.

\end{lstlisting}

\end{document}